\documentclass[runningheads]{llncs}
\usepackage[T1]{fontenc}
\usepackage{graphicx}
\usepackage{comment}
\usepackage{amsmath,amssymb} 
\usepackage{epsfig}
\usepackage{graphicx}
\usepackage{amsmath}
\usepackage{amssymb}
\usepackage{microtype}
\usepackage{mwe}
\usepackage{wrapfig}
\usepackage{cite}

\usepackage[usenames, dvipsnames]{color}
\usepackage[ruled]{algorithm2e}
\usepackage[noend]{algpseudocode}
\usepackage{wrapfig}
\usepackage{tablefootnote}
\usepackage{paralist}  
\usepackage[export]{adjustbox}
\usepackage{array}
\usepackage{pbox}
\usepackage{multirow}
\usepackage{booktabs}
\usepackage{soul}
\usepackage{subcaption}
\usepackage{enumitem}
\setitemize{noitemsep,topsep=0pt,parsep=0pt,partopsep=0pt}
\usepackage{xcolor}
\usepackage{cite}

\newcommand\etal{\textit{et al.}}
\newcommand\ie{\textit{i.e. }}
\newcommand\eg{\textit{e.g.}}

\newcolumntype{L}[1]{>{\raggedright\let\newline\\\arraybackslash\hspace{0pt}}m{#1}}
\newcolumntype{C}[1]{>{\centering\let\newline\\\arraybackslash\hspace{0pt}}m{#1}}
\newcolumntype{R}[1]{>{\raggedleft\let\newline\\\arraybackslash\hspace{0pt}}m{#1}}

\definecolor{citecolor}{HTML}{0071bc}

\usepackage[pagebackref=true,breaklinks=true,letterpaper=true,colorlinks,citecolor=citecolor,bookmarks=false]{hyperref}

\begin{document}


\pagestyle{headings}
\mainmatter
\def\ECCVSubNumber{4531}  

\title{
Improving Face Recognition by Clustering\\
Unlabeled Faces in the Wild
}

\titlerunning{Improving Face Recognition by Clustering Unlabeled Faces in the Wild}
%
\author{Aruni RoyChowdhury\inst{1}\thanks{Now at Amazon, work done prior to joining, while interning at NEC Labs America.} \and
Xiang Yu\inst{2} \and
Kihyuk Sohn\inst{2}\thanks{Now at Google, work done prior to joining.} \and \\
Erik Learned-Miller\inst{1} \and
Manmohan Chandraker \inst{2,3}
}
\authorrunning{A. RoyChowdhury et al.}
%
\institute{University of Massachusetts, Amherst \and
NEC Labs America \and
University of California, San Diego
}

\maketitle
\thispagestyle{empty}

\begin{abstract}
    
While deep face recognition has benefited significantly from large-scale labeled data, current research is focused on leveraging unlabeled data to further boost performance, reducing the cost of human annotation. Prior work has mostly been in controlled settings, where the labeled and unlabeled data sets have no overlapping identities by construction. This is not realistic in large-scale face recognition, where one must contend with such overlaps, the frequency of which increases with the volume of data. Ignoring identity overlap leads to significant labeling noise, as data from the same identity is split into multiple clusters. To address this, we propose a novel identity separation method based on extreme value theory. It is formulated as an out-of-distribution detection algorithm, and greatly reduces the problems caused by overlapping-identity label noise.  Considering cluster assignments as pseudo-labels, we must also overcome the labeling noise from clustering errors. We propose a modulation of the cosine loss, where the modulation weights correspond to an estimate of clustering uncertainty. 
Extensive experiments on both controlled and real settings demonstrate our method's consistent improvements over supervised baselines, e.g., 11.6\% improvement on IJB-A verification.

\end{abstract}


\section{Introduction}
\label{sec:intro}

Deep face recognition has achieved impressive  performance, benefiting from large-scale labeled data. Examples include DeepFace~\cite{deepface}, which uses 4M labeled faces for training and FaceNet~\cite{schroff2015facenet}, which is trained on 200M labeled faces. Further improvements in recognition performance using traditional supervised learning may require tremendous annotation efforts to increase the labeled dataset volume, which is impractical, labor intensive and does not scale well. Therefore, exploiting unlabeled data to augment the labeled data, i.e., {\em semi-supervised learning}, is an attractive alternative. Preliminary work on generating pseudo-labels by clustering unlabeled faces has been shown to be effective in improving performance under controlled settings~\cite{sohn2018unsupervised,zhan2018consensus,yang2019learning}. 
%

\begin{figure}[!!t]
    \centering
    \includegraphics[width=0.85\textwidth]{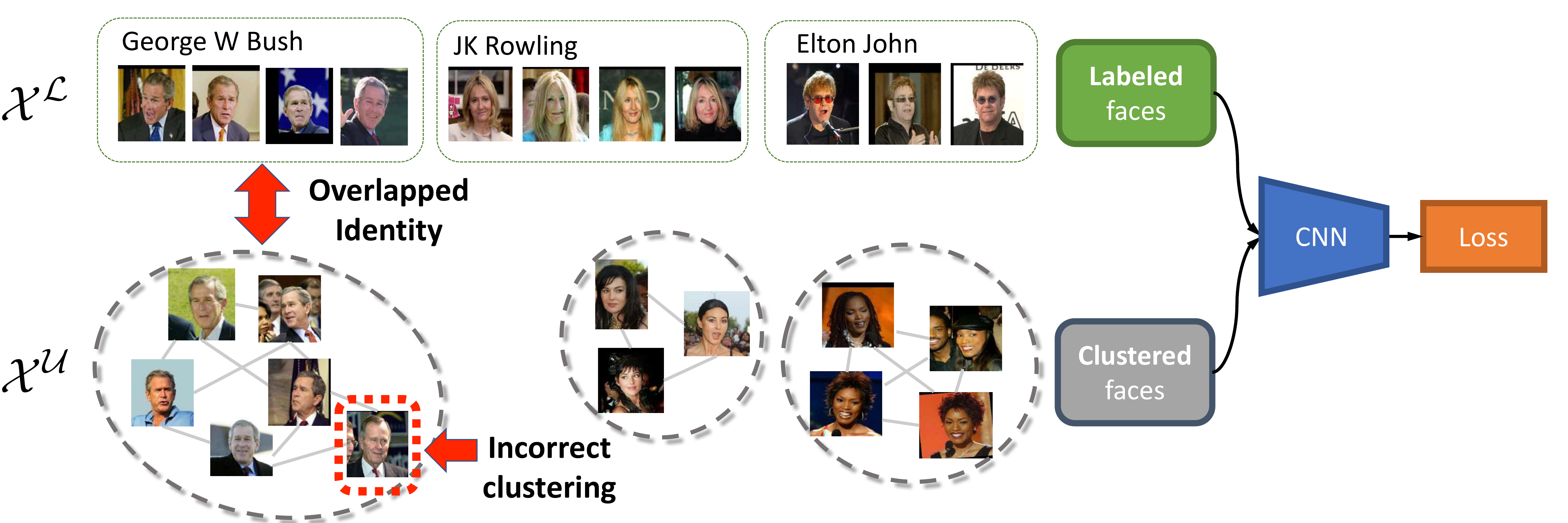}
    \caption{ 
        \small{
        Given a face recognition model trained on labeled faces ($\mathcal{X^L}$), we wish to cluster unlabeled data ($\mathcal{X^U}$) for additional training samples to further improve recognition performance. Key challenges include \textit{\textbf{overlapping identities}} between labeled and unlabeled data (\textit{George W Bush} images present in both $\mathcal{X^L}$ and $\mathcal{X^U}$) as well as noisy training labels arising from \textit{\textbf{incorrect cluster assignments}} (a picture of \textit{George Bush Sr.} is erroneously assigned to a cluster of \textit{George W Bush} images).
        } }
    \label{fig:overlap-teaser}
    \vspace{-0.6cm}
\end{figure}

However, although learning from unlabeled data is a mature area and theoretically attractive, face recognition as a field has yet to adopt such methods in practical and realistic settings. 
There are several obstacles to directly applying such techniques that are peculiar to the setting of large-scale face recognition. First, there is a common assumption of semi-supervised face recognition methods is that there is \textit{no class or identity overlap} between the unlabeled and the labeled data. This seemingly mild assumption, however, violates the basic premise of semi-supervised learning -- that nothing is known about the labels of the unlabeled set.  
Thus, either practitioners must manually verify this property, meaning that the data is no longer \textit{``truly'' unlabeled}, or proceed under the assumption that identities are disjoint between labeled and unlabeled training datasets, which inevitably introduces labeling noise. When such overlapping identities are in fact present (Fig.~\ref{fig:overlap-teaser}), a significant price is paid in terms of performance, as demonstrated empirically in this work.
A further practical concern in face recognition is the availability of massive labeled face datasets. 
Most current work using unlabeled faces focus on improving the performance of models trained with \textit{limited labeled data}~\cite{zhan2018consensus,yang2019learning}, and it is unclear if there are any benefits from using unlabeled datasets when baseline face recognition models are trained on \textit{large-scale labeled data}.

In this paper, we present recipes for exploiting unlabeled data to further improve the performance of fully supervised state-of-the-art face recognition models, which are mostly trained on large-scale labeled datasets. 
%
We demonstrate that learning from unlabeled faces is indeed a practical avenue for improving deep face recognition, also addressing important practical challenges in the process -- accounting for overlapping identities between labeled and unlabeled data, and attenuating the effect of noisy labels when training on pseudo-labeled data.

\begin{wrapfigure}{r}{0.5\textwidth}
    \centering
    \vspace{-2mm}
    \vspace{-0.3cm}
    \includegraphics[width=0.5\textwidth]{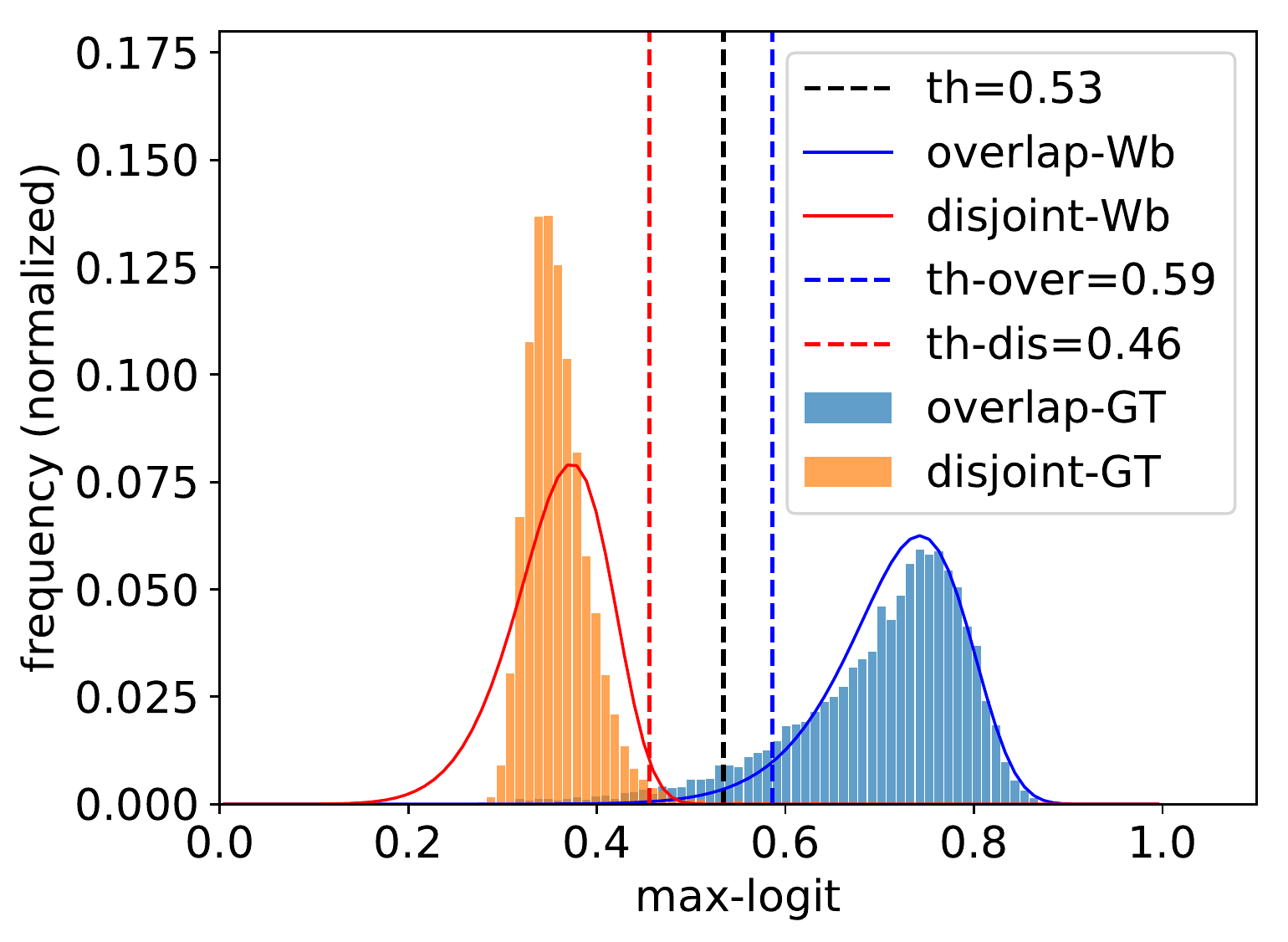}
    \vspace{-0.8cm}
    \caption{ \small{
    Extreme value theory (EVT) provides a principled way of setting thresholds on the max-logits $\boldsymbol{z}_i$ for $\mathbf{x}_i \in \mathcal{X^U}$ to separate \textit{disjoint} and \textit{overlapping} identities ({\color{red}{red}} and {\color{blue}blue} vertical lines).
    An initial threshold is determined by Otsu's method (black vertical line). 
    This plot uses splits from the MS-Celeb-1M dataset~\cite{guo2016msceleb}.
      } } 
    \label{fig:class-overlap}
    \vspace{-0.7cm}
\end{wrapfigure}

We begin with Face-GCN \cite{yang2019learning}, a graph convolutional neural network (GCN) based face clustering method, to obtain pseudo-labels on unlabeled faces. To deal with the overlapping identity problem, we observe that the distribution of classification confidence on overlapping and disjoint identities is different -- since our initial face feature is provided by a recognition engine trained on known identities, the confidence score of the overlapping identity images should be higher than those of non-overlapping identity images, as visualized in Fig.~\ref{fig:class-overlap}. 
Based on this observation, we approach the problem as \textit{out-of-distribution detection}~\cite{hendrycks2016baseline,liang2017enhancing,lee2018simple}, and propose to parameterize the distribution of confidence scores as a mixture of Weibulls, motivated by extreme value theory. This results in an unsupervised procedure to separate overlapping identity samples from unlabeled data on-the-fly.

After resolving the overlapping identity caused label noise, the \textit{systematic label noise} from the clustering algorithm remains, which is another prime cause for deteriorating performance in face recognition~\cite{wang2018devil}. 
Instead of an additional complicated pruning step to discard noisy samples, \eg~as done in \cite{yang2019learning}, we deal with the label noise during the re-training loop using the joint data of both labeled and clustered faces, by introducing a simple clustering uncertainty based attenuation on the training loss to reduce the effect of erroneous gradients caused by the noisy labeled data. This effectively smoothes the re-training procedure and has shown clear performance gains in our experiments.
Our contributions are summarized as the following:
\begin{itemize}
    \item{To our knowledge, we are the first to tackle the practical issue of overlapping identities between labeled and unlabeled face data during clustering, formulated as an out-of-distribution detection.}
    
    \item{We successfully demonstrate that jointly leveraging large scale unlabeled data along with labeled data in a semi-supervised fashion can indeed significantly improve over supervised face recognition performance, i.e., substantial gains over a supervised CosFace~\cite{wang2018cosface} model across multiple public benchmarks.}
    
    \item{We introduce a simple and scalable uncertainty-modulated training loss into the semi-supervised learning setup, which is designed to compensate for the label noise introduced by the clustering procedure on unlabeled data.}
    
    \item{We provide extensive and ablative insights on both controlled and real-world settings, serving as a recipe for the semi-supervised face recognition or other large scale recognition problems.}
    
\end{itemize}


\vspace{-2mm}
\section{Related Work}
\label{sec:related}

\textbf{Face Clustering:} Jain~\cite{jain2010data} provides a survey on classic clustering techniques. Most recent approaches~\cite{otto2017clustering,lin2017proximity,shi2018face,lin2018deep,erdosrenyi} work on face features extracted from supervisedly-trained recognition engines.
``Consensus-driven propagation'' (CDP)~\cite{zhan2018consensus} assigns pseudo-labels to unlabeled faces by forming a graph over the unlabeled samples. An ensemble of various network architectures provides multiple views of the unlabeled data, and an aggregation module decides on positive and negative pairs. 
Face-GCN~\cite{yang2019learning} formulates the face clustering problem into a regression for cluster proposal purity, which can be fully supervised.
Re-training the recognition engine with the clustered ``pseudo-identities''  and the original data improves performance, however, CDP~\cite{zhan2018consensus} and Face-GCN re-training assumes the ``pseudo-identities" and the original identities have no overlap, which does not always hold true. Meanwhile, their investigation stays in a controlled within-distribution setting using the MS-Celeb-1M dataset~\cite{guo2016msceleb}, which is far from realistic. In contrast, we demonstrate that these considerations are crucial to achieve gains for practical face recognition with truly large-scale labeled and unlabeled datasets.

\noindent\textbf{Out-Of-Distribution Detection:} Extreme value distributions have been used in calibrating classification scores~\cite{rudd2017extreme}, classifier meta-analysis~\cite{scheirer2011meta}, open set recognition robust to adversarial images~\cite{bendale2016towards} and as normalization for score fusion from multiple biometric models~\cite{scheirer2010robust}, which is quite different from our problem.
Recent approaches to out-of-distribution detection utilize the confidence of the predicted posteriors~\cite{hendrycks2016baseline,liang2017enhancing}, while Lee~\etal~\cite{lee2018simple} use Mahalanobis distance-based classification along with gradient-based input perturbations. \cite{lee2018simple} outperforms the others, but does not scale to our setting -- estimating per-subject covariance matrices is not feasible for the typical long-tailed class distribution in face recognition datasets. 
%

\noindent\textbf{Learning with Label Noise:} Label-noise~\cite{natarajan2013learning} has a significant effect on the performance of the face embeddings obtained from face recognition models trained on large datasets, as extensively studied in Wang~\etal~\cite{wang2018devil}. Indeed, even large scale human-annotated face datasets such as the well-known MS 1 Million (MS-1M) are shown to have some incorrect labeling, and gains in recognition performance can be attained by cleaning up the labeling~\cite{wang2018devil}.
%
%
Applying label-noise modeling to our problem of large-scale face recognition has its challenges --  the labeled and unlabeled datasets are class-disjoint, a situation not considered by earlier methods~\cite{patrini2017making,li2017learning,hendrycks2018using}; having $\sim$100k identities, typically long-tailed, make learning a label-transition matrix challenging~\cite{patrini2017making,hendrycks2018using}; label-noise from clustering pseudo-labels is typically structured and quickly memorized by a deep network, unlike the uniform-noise experiments in \cite{ICML2019_UnsupervisedLabelNoise,zhang2017mixup,Forgetting}. Our unsupervised label-noise estimation does not require a clean labeled dataset to learn a training curriculum unlike \cite{jiang2017mentornet,ren2018learning}, and can thus be applied out-of-the-box. 
%

\vspace{-2mm}
\section{Learning from Unlabeled Faces}
\label{sec:method}


\begin{figure*}[!!t]
    \centering
    \includegraphics[width=0.95\textwidth]{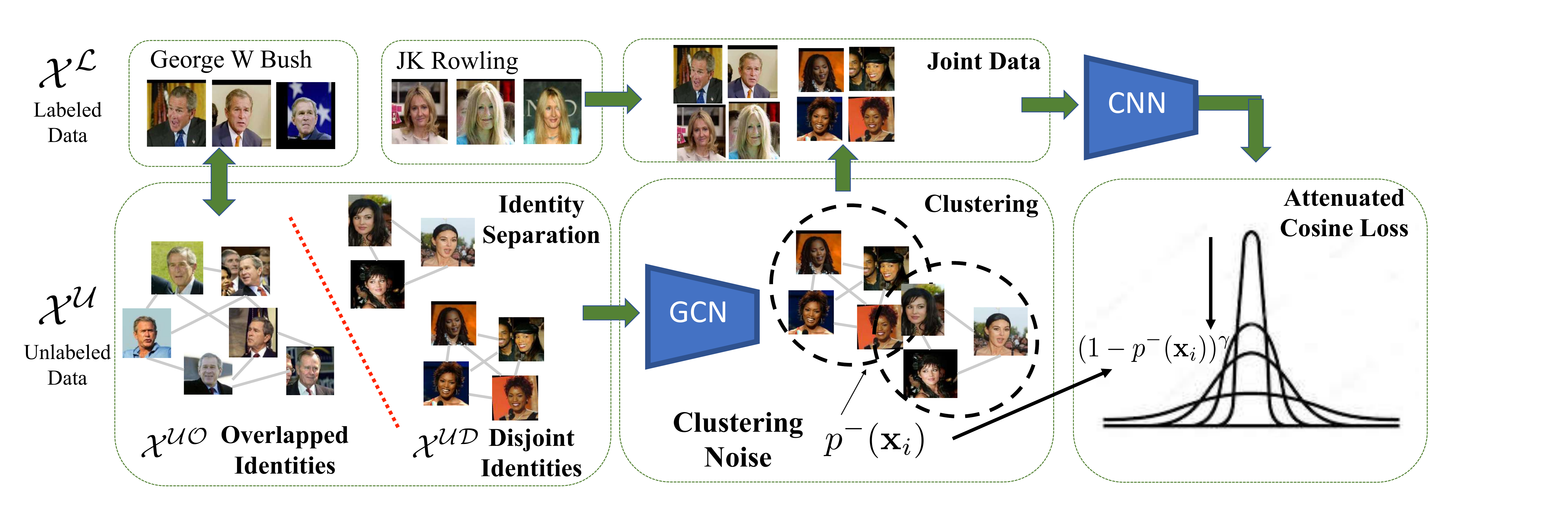}
    \caption{ \small{
        Our approach trains a deep neural network~\cite{wang2018cosface} jointly on labeled faces $\mathcal{X^L}$ and unlabeled faces $\mathcal{X^U}$.
        Unlabeled samples with \textbf{\textit{overlapping}} and \textbf{\textit{disjoint}} identities w.r.t. $\mathcal{X^L}$ are separated into  $\mathcal{X^{UO}}$ and $\mathcal{X^{UD}}$, respectively (Sec.~\ref{sec:overlap}).
        The unlabeled faces in $\mathcal{X^{UD}}$ are clustered using a \textbf{\textit{graph conv-net}} or \textbf{\textit{GCN}} (Sec.~\ref{sec:gcn}).
        Estimates of \textbf{\textit{cluster uncertainty}} $p^-(\mathbf{x}_i)$ are used to modulate the cosine loss during re-training (Sec.~\ref{sec:retrain}).
        }}
    \label{fig:overview}
    \vspace{-0.5cm}
\end{figure*}


Formally, let us consider samples $\mathcal{X} = \{ \mathbf{x}_i \}_{i \in [n]}$, divided into two parts: $\mathcal{X^L}$ and  $\mathcal{X^U}$ of sizes $l$ and $u$ respectively.
Now $\mathcal{X^L} := \{\mathbf{x}_1, ... , \mathbf{x}_l  \}$ consist of faces that are provided with identity labels $\mathcal{Y^L} := \{y_1, ... , y_l  \}$, while we do not know the identities of the unlabeled faces $\mathcal{X^U} := \{\mathbf{x}_{l+1}, ... , \mathbf{x}_{l+u}  \}$.
Our approach aims to improve the performance of a supervised face recognition model, trained on ($\mathcal{X^L}$, $\mathcal{Y^L}$), by first clustering the unlabeled faces $\mathcal{X^U}$, then re-training on both labeled and unlabeled faces, using the cluster assignments on $\mathcal{X^U}$ as pseudo-labels.
Fig.~\ref{fig:overview} visually summarizes the steps -- \textbf{(1)}~train a supervised face recognition model on ($\mathcal{X^L}$, $\mathcal{Y^L}$); \textbf{(2)}~separate the samples in $\mathcal{X^U}$ having overlapping identities with the labeled training set; \textbf{(3)}~cluster the disjoint-identity unlabeled faces; \textbf{(4)}~learn an unsupervised model for the likelihood of incorrect cluster assignments on the pseudo-labeled data; \textbf{(5)}~re-train the face recognition model on labeled and pseudo-labeled faces, attenuating the training loss for pseudo-labeled samples using the estimated clustering uncertainty. 
In this section, we first describe the separation of overlapping identity samples from unlabeled data (Sec.~\ref{sec:overlap}), followed by an overview of the face clustering procedure (Sec.~\ref{sec:gcn}) and finally re-training the recognition model with an estimate of clustering uncertainty (Sec.~\ref{sec:retrain}).

\subsection{Separating Overlapping Identities}
\label{sec:overlap}

\textbf{Overlapping identities.} We typically have no control over the gathering of the unlabeled data $\mathcal{X^U}$, so the same subject {\em{S}} may exist in labeled data (thus, be a class on which the baseline face recognition engine is trained) and also within our unlabeled dataset, \ie $\mathcal{X^U} = \mathcal{X^{UO}} \cup \mathcal{X^{UD}}$, where $\mathcal{X^{UO}}$ and $\mathcal{X^{UD}}$ denote the identity overlapped and identity disjoint subsets of $\mathcal{X^U}$.
%
By default, the clustering will assign images of subject {\em{S}} in the unlabeled data as a new category. In this case, upon re-training with the additional pseudo-labeled data, the network will incorrectly learn to classify images of subject {\em{S}} into \textit{two} categories. 
This is an important issue, since overlapping subjects can occur naturally in datasets collected from the Internet
or recorded through passively mounted cameras, which to our knowledge has not been directly addressed by most recent pseudo-labeling methods~\cite{zhan2018consensus,sohn2018unsupervised,yang2019learning}. 

\noindent
\textbf{Out-of-distribution detection.} The problem of separating unlabeled data into samples of disjoint and overlapping classes (w.r.t. the classes in the labeled data) can be regarded as an ``out-of-distribution'' detection problem. 
The intuition is that unlabeled samples with overlapping identities will have higher confidence scores from a face recognition engine, as the same labeled data is used to train the recognition model~\cite{hendrycks2016baseline}.
Therefore, we search for thresholds on the recognition confidence scores that can separate disjoint and overlapping identity samples.
Note, since the softmax operation over several thousand categories can result in small values due to normalization, we use the \textit{maximum logit} $\boldsymbol{z}_i$ for each sample $\mathbf{x}_i \in \mathcal{X^U}$ as its confidence score.
Since the $\boldsymbol{z}_i$ are the maxima over a large number of classes, we can draw upon results from extreme value theory (EVT) which state that the limiting distribution of the maxima of i.i.d random variables belongs to either the Gumbel, Fr\'echet or Weibull family~\cite{de2007extreme}. Specifically, we model the $\boldsymbol{z}_i$ using the Weibull distribution,
\begin{equation}
    f(\boldsymbol{z}_i;\lambda,k) =
        \begin{cases}
        \frac{k}{\lambda}\left(\frac{\boldsymbol{z}_i}{\lambda}\right)^{k-1}e^{-(\boldsymbol{z}_i/\lambda)^{k}} & \boldsymbol{z}_i\geq0 ,\\
        0 & \boldsymbol{z}_i<0,
        \end{cases}
\end{equation}
where $k>0$ and $\lambda>0$ denote the shape and scale parameters, respectively. 
We use Otsu's method~\cite{otsu1979threshold} to obtain an initial threshold on the $\boldsymbol{z}_i$ values, then fit a two-component mixture of Weibulls, modeling the \textit{identity-overlapping} and \textit{identity-disjoint} sets $\mathcal{X^{UO}}$ and $\mathcal{X^{UD}}$, respectively.
Selecting values corresponding to 95\% confidence under each Weibull model provides thresholds for deciding if $\mathbf{x}_i \in \mathcal{X^{UO}}$ or $\mathbf{x}_i \in \mathcal{X^{UD}}$ with high confidence; we reject samples that fall outside of this interval.
This approach does not require setting any hyper-parameters a priori, and can be applied to any new unlabeled dataset.

\subsection{Clustering Faces with GCN}
\label{sec:gcn}
We use Face-GCN \cite{yang2019learning} to assign pseudo-labels for unlabeled faces in $\mathcal{X^{UD}}$, which leverages a graph convolutional network (GCN)~\cite{kipf2016semi} for large-scale face clustering.
We provide a brief overview of the approach for completeness. 
%
%
Based on features extracted from a pre-trained face recognition engine, a nearest-neighbor graph is constructed over all samples.
By setting various thresholds on the edge weights of this graph, a set of connected components or cluster proposals are generated. 
During training, the aim is to regress the precision and recall of the cluster proposals arising from a single ground truth identity, motivated by object detection frameworks~\cite{maskrcnn}.
Since the proposals are generated based on labeled data, the Face-GCN is trained in a fully supervised way, unlike regular GCN training, which are typically trained with a classification loss, either for each node or an input graph as a whole.
During testing, a ``de-overlap'' procedure uses predicted GCN scores for the proposals to partition an unlabeled dataset into a set of clusters. Please see \cite{yang2019learning} for further details.

\subsection{Joint Data Re-training with Clustering Uncertainty}
\label{sec:retrain}

We seek to incorporate the uncertainty of whether a pseudo-labeled (\ie clustered) sample was correctly labeled into the face recognition model re-training.
Let a face drawn from the unlabeled dataset  be $\mathbf{x}_i \in \mathcal{X^{UD}}$. The feature representation for that face using the baseline supervised model is denoted as $\Phi(\mathbf{x}_i)$. Let cluster assignments obtained on $\mathcal{X^{UD}}$ be $\{ \mathcal{C}_1, \mathcal{C}_2, ... ,  \mathcal{C}_K \}$, for $K$ clusters. We train a logistic regression classifier to estimate $P(\mathcal{C}_k \mid \Phi(\mathbf{x}_i))$, for $k = 1,2, ... K$,
\begin{align}
    P(\mathcal{C}_k \mid \Phi(\mathbf{x}_i)) =  \frac{\exp(\mathbf{\omega}_k^{\top}\Phi(\mathbf{x}_i))}{ \sum_{j} \exp( \mathbf{\omega}_j^{\top}\Phi(\mathbf{x}_i))}
    \label{eq:logistic}
\end{align}
where $\mathbf{\omega}_k$ are the classifier weights for the $k$-th cluster. 
Intuitively, we wish to determine how well a simple linear classifier on top of discriminative face descriptors can fit the cluster assignments.
We compare the following uncertainty metrics: 
\textbf{(1)}~\textit{Entropy} of the posteriors across the $K$ clusters, \ie $\sum_k P(\mathcal{C}_k \mid \Phi(\mathbf{x}_i)) \log P(\mathcal{C}_k \mid \Phi(\mathbf{x}_i))$; 
\textbf{(2)}~\textit{Max-logit:} the largest logit value over the $K$ clusters, 
%
\textbf{(3)}~\textit{Classification margin:} difference between the max and the second-max logit, indicating how easily a sample can flip between two clusters. 

\begin{figure}[t]
    \centering
    \begin{tabular}{ccc}
      \includegraphics[width=0.30\textwidth,trim=0 0 0 1cm, clip]{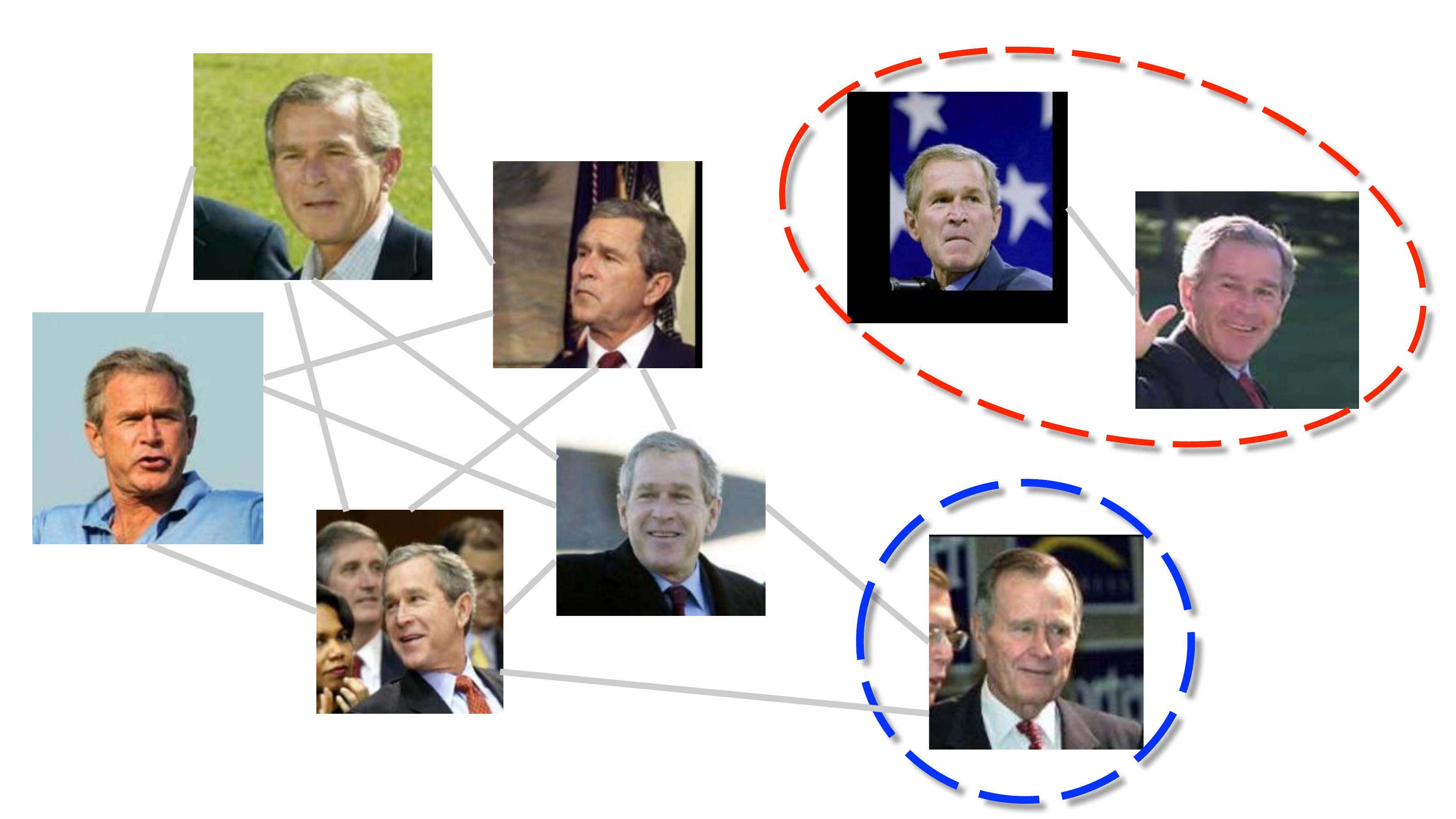} & 
      \includegraphics[width=0.27\textwidth]{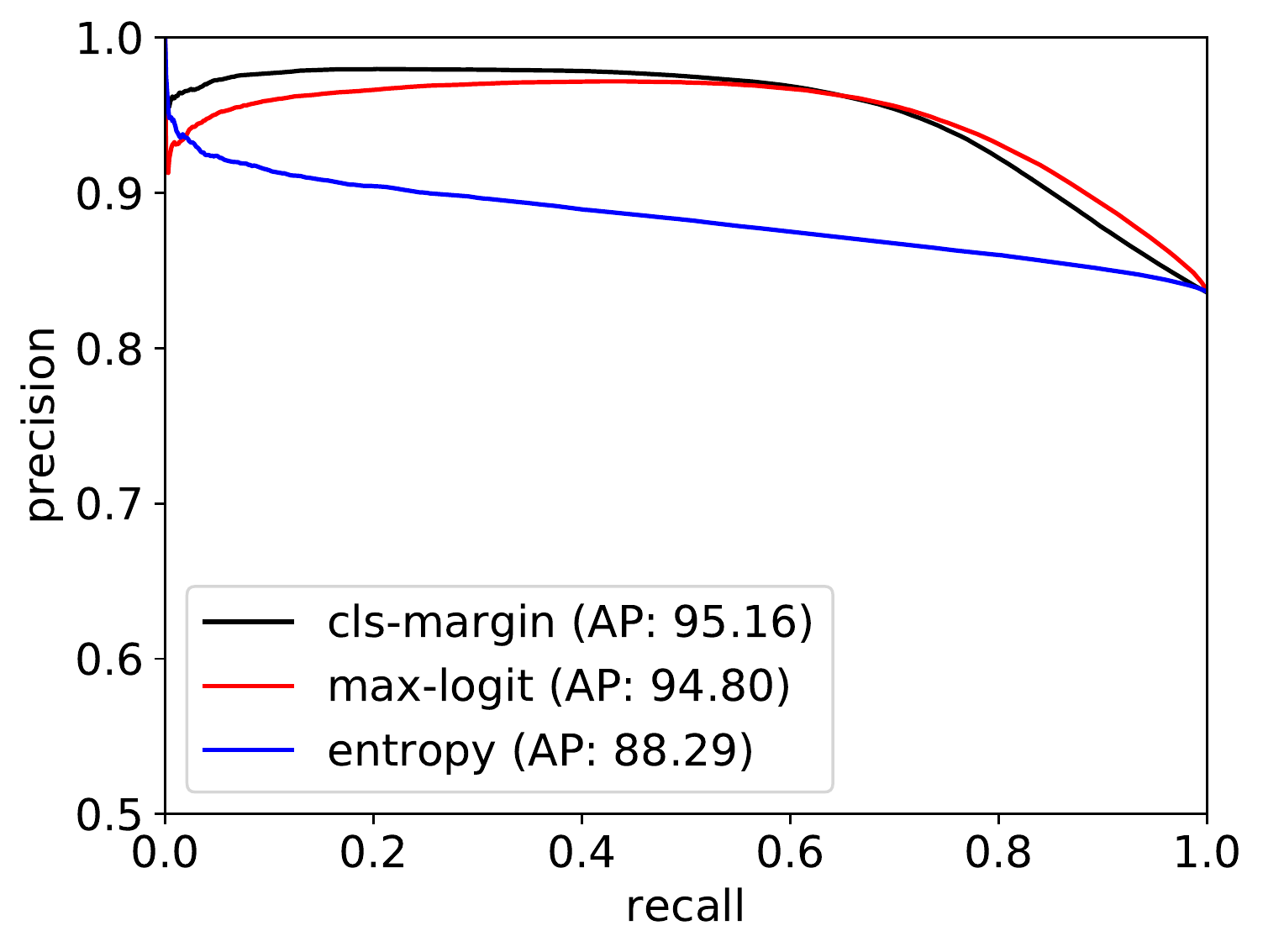}  &
      \includegraphics[width=0.28\textwidth]{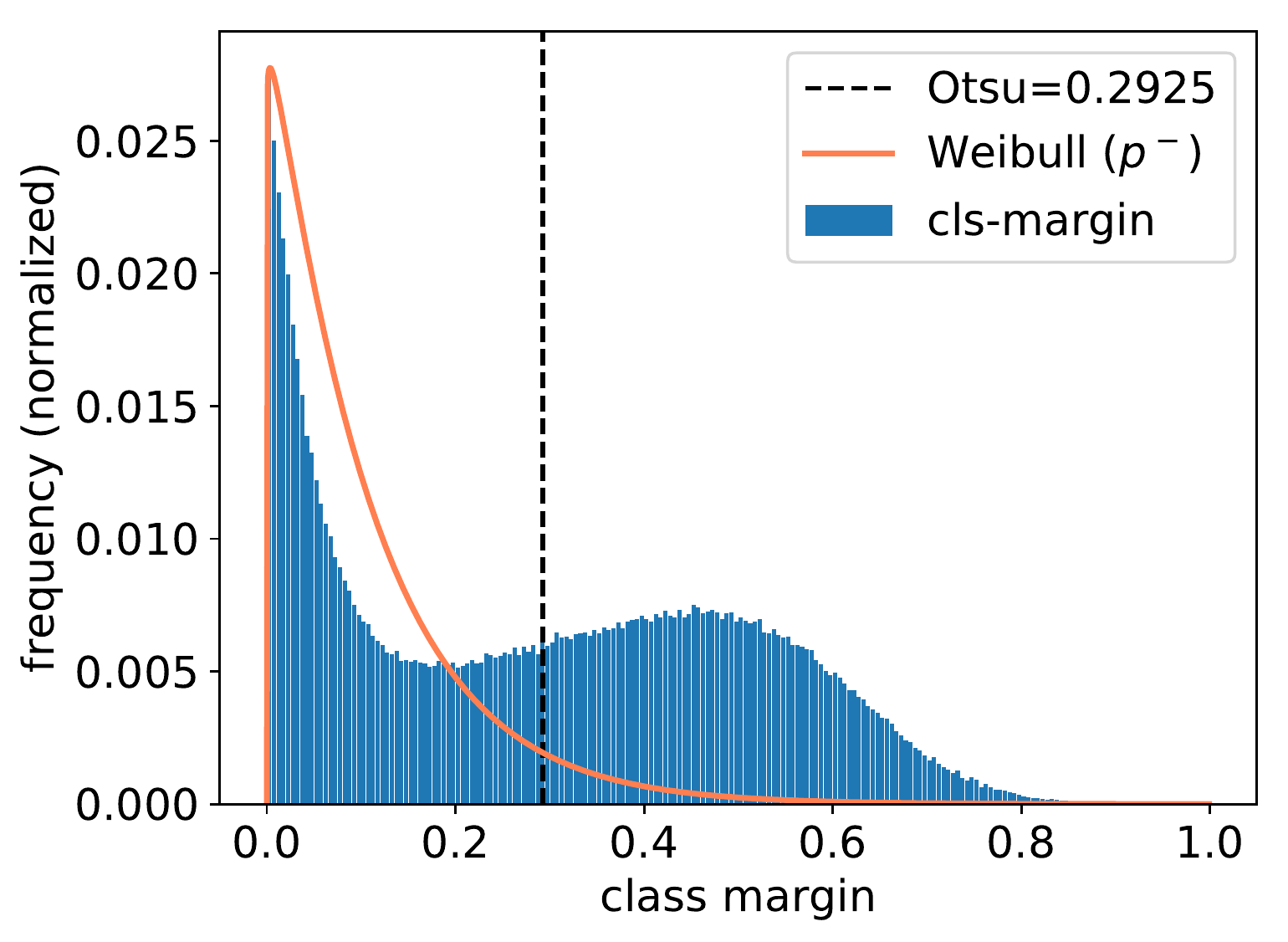} \\
     (a) & (b) & (c) \\
    \end{tabular}
    \caption{ \small{ \textbf{Clustering uncertainty.}
    \textbf{(a)} Illustration of incorrect pseudo-labels -- an image of \textit{George Bush Sr.} is included in a cluster of \textit{George W Bush} images (outlier circled in {\color{blue}blue}); some \textit{George W Bush} images are spread across multiple clusters (``split ID'' circled in {\color{red}red}).
    \textbf{(b)} Precision-recall curves showing Average Precision (AP) of predicting if a cluster assignment is correct using class-margin, max-logit and entropy. 
    \textbf{(c)} Estimating clustering error $p^-(\mathbf{x}_i)$ from the distribution of class-margin ({\color{BurntOrange}orange} curve).
    }
    } 
    \label{fig:soft-score}
\end{figure}

\noindent
We consider two kinds of incorrect clustering corresponding to notions of precision and recall: 
\textbf{(1)} \textbf{Outliers}, samples whose identity does not belong to the identity of the cluster; \textbf{(2)} \textbf{Split-ID}, where samples from the same identity are spread over several clusters (Fig.~\ref{fig:soft-score}(a)). In a controlled setting with known ground-truth identities, we validate our hypothesis that the uncertainty measures can distinguish between correct and incorrect cluster assignments (Fig.~\ref{fig:soft-score}(b)). Note that Split-ID makes up the bulk of incorrectly-clustered samples, while outliers are about 10\%. 
%
%
\noindent
Fig.~\ref{fig:soft-score}~(c) shows the distribution of class-margin on pseudo-labeled data on one split of the MS-1M dataset. Intuitively, samples that do not have a large classification margin are likely to be incorrect pseudo-labels,
resulting in a bi-modal distribution -- noisily labeled samples in one mode, and correctly labeled samples in the other.
Notice that similar to overlapping v.s. disjoint identity, this is another distribution separation problem.
A Weibull is fit to the lower portion of the distribution (orange curve), with an initial mode-separating threshold obtained from Otsu's method (black vertical line). 
The probability of sample $\mathbf{x}_i$ being incorrectly clustered is estimated by:
\begin{align}
    p^-(\mathbf{x}_i) = P(g(\mathbf{x}_i) \mid \theta_{Wb}^-),
    \label{eq:distribution_fit}
\end{align}
where $\theta_{Wb}^-$ are the parameters of the learned Weibull model, $g(.)$ denotes the measure of uncertainty, \eg class-margin. 
Note, ground-truth labels are not required for this estimation.
We propose to associate the above uncertainty with the pseudo-labeled samples and set up a probabilistic face recognition loss.

\noindent
The large margin cosine loss~\cite{wang2018cosface} is used for training:
\begin{align}
    \mathcal{L}(\mathbf{x}_i)=-\log \frac{\exp(\alpha (\mathbf{w}_j^{\top}\mathbf{f}_i-m))}{\exp(\alpha (\mathbf{w}_j^{\top}\mathbf{f}_i-m)) + \sum_{k\neq j} \exp(\alpha \mathbf{w}_k^{\top}\mathbf{f}_i)}
    \label{eq:cosloss}
\end{align}
where $\mathbf{f}_i$ is the deep feature representation of the $i$-th training sample $\mathbf{x}_i$, $\mathbf{w}_j$ is the learned classifier weight for the $j$-th class, $m\in[0,1]$ is an additive margin and $\alpha$ is a scaling factor;
 $\Vert\mathbf{f}_i\Vert$ and $\Vert\mathbf{w}_j\Vert$ are set to $1$. For $\mathbf{x}_i \in \mathcal{X^U}$, we modulate the training loss with the clustering uncertainty $p^-(\mathbf{x}_i)$, where $\gamma$ controls the weighting curve shape:
\begin{align}
    \mathcal{L}^{p}(\mathbf{x}_i) = (1-p^-(\mathbf{x}_i))^{\gamma}\mathcal{L}(\mathbf{x}_i),
    \label{eq:softlabel}
\end{align}

\vspace{-2mm}
\section{Experiments}
\label{sec:experiments}

\vspace{-3mm}
We augment supervised models trained on labeled data with additional pseudo-labeled data under various scenarios. We summarize the main findings first --- 
{\bf{\em(i)}}~the baseline supervised model benefits from additional pseudo-labeled training data;
{\bf{\em(ii)}}~re-training on clustering without handling overlapping IDs can hurt performance, and our approach of separating overlaps is shown to be effective empirically;
{\bf{\em(iii)}}~increasing diversity of training data by using unlabeled data from outside the distribution of the labeled set helps more than comparable amounts of within-domain unlabeled data; 
{\bf{\em(iv)}}~scaling up to using the entire MS-Celeb-1M~\cite{guo2016msceleb} dataset (or MS1M for short) as labeled training set, as typically done by most deep face models, we see significant gains in performance \textit{only} when the volume of unlabeled samples is comparable to the size of MS1M itself.

\noindent
{\bf Experimental setup.}
\label{sec:exp_setup}
Table~\ref{tab:train_data} summarizes the  training data sources. The cleaned version of MS1M dataset contains 84,247 identities and 4,758,734 samples in total. Partitioning on the identities, the full MS1M dataset is split into 10 parts with approximately 8.4k identities and 470k samples per split. We create the following settings:
\begin{itemize}
    \item{\textit{\textbf{Controlled disjoint} (Sec.~\ref{sec:ms1m-control}):} Both labeled and unlabeled data are drawn from splits of MS1M (Table~\ref{tab:train_data} \textit{MS1M} splits 1 and 2, respectively). Thus, they have the same distribution and have no overlapping identities by construction, similar to the setting in \cite{yang2019learning}. We compare baseline clustering methods and the effect of clustering uncertainty on re-training the face recognition model.} 
    
    \item{\textit{\textbf{Controlled overlap} (Sec.~\ref{sec:id-overlap}):} we introduce simulated identity overlap between the two datasets (Table~\ref{tab:train_data} \textit{MS-Celeb-1M} splits \textit{1-O} and \textit{2-O}), showing the detrimental effect of na\"{i}vely clustering and re-training in this case, and the efficacy of our proposed approach.}
    
    \item{\textit{\textbf{Semi-controlled} (Sec.~\ref{sec:semi}):} we have limited labeled data (split-1 of MS1M) with unlabeled data from another dataset, i.e., VGGFace2~\cite{cao2018vggface2}, containing 8.6k identities and 3.1 million images. This is closer to the realistic scenario, with potential identity overlaps and distribution shift between data sources.}
    
    \item{\textit{\textbf{Uncontrolled} (Sec.~\ref{sec:uncontrol_vgg}):} close to the real-world setting, we use \textit{all} the labeled data at our disposal (entire MS-Celeb-1M) and try to improve performance further by including unlabeled data from other datasets -- VGGFace2~\cite{cao2018vggface2}, IMDB-SenseTime~\cite{wang2018devil}, CASIA~\cite{yi2014learning} \& GlintAsian~\cite{glintasian}, by completely ignoring their ground truth labels. Note, this setting is not addressed in prior art on pseudo-labeling faces~\cite{zhan2018consensus,yang2019learning}.}
\end{itemize}

\noindent
\textbf{Evaluation.}
We report results on the following: verification accuracy on \textit{Labeled Faces in the Wild} (LFW)~\cite{huang2008labeled,learned2016labeled} and \textit{Celebrity Frontal to Profile} (CFP)~\cite{sengupta2016frontal}; identification at rank-1 and rank-5, and True Accept Rate (TAR) at False Accept Rates (FAR) of 1e-3 and 1e-4 on the challenging \textit{IARPA Janus} benchmark (IJB-A)~\cite{klare2015pushing}.
For clustering metrics we adopt the protocol used in \cite{yang2019learning}.

\noindent
\textbf{Training details.} 
We use the high-performing CosFace model~\cite{wang2018cosface} for face recognition, with a 118-layer ResNet backbone, trained for 30 epochs on labeled data. Re-training is done from scratch with identical settings. Face-GCN uses the publicly available code of GCN-D~\cite{yang2019learning}. For further details please refer to the supplementary materials.

\begin{table}[t]
\parbox{.45\linewidth}{
\centering
\small
\caption{\small{Statistics for training datasets.}}
\label{tab:train_data}
 \begin{tabular}{@{\extracolsep{2pt}}lrr}
 \toprule
  \textbf{Dataset}    & \textbf{\#IDs}  & \textbf{Images} \\
 \midrule 
  MS-Celeb-full         & 84k    & 4.7M  \\
  MS-Celeb-split-1      & 8.4k      & 505k    \\
  MS-Celeb-split-2      & 8.4k      & 467k  \\
  MS-Celeb-split-1-O    & 16.8k      & 729k    \\
  MS-Celeb-split-2-O    & 16.8k      & 705k  \\
 \midrule
  VGGFace2~\cite{cao2018vggface2}        & 8.6k     & 3.1M \\
  CASIA-WebFace~\cite{yi2014learning}    & 10.5k    & 455k \\
  IMDB-SenseTime~\cite{wang2018devil}    & 51k    & 1M \\
  GlintAsian~\cite{glintasian}           & 94k    & 2.8M \\
 \bottomrule
 \end{tabular}%
}
\hfill
\parbox{.45\linewidth}{
\centering
\small
\caption{\small{\textbf{Controlled: Face clustering baselines.} Comparing the GCN-based method with standard clustering algorithms. The GCN is trained on MS-Celeb-1M split 1, tested on split 2.}}
\label{tab:cluster_ms1m}
\begin{tabular}{@{\hskip 1.5mm}l@{\hskip 1.5mm}c@{\hskip 1.5mm}c@{\hskip 1.5mm}c@{\hskip 1.5mm}r}
\toprule
 \textbf{Method}  & \textbf{Prec}  & \textbf{Rec}  & \textbf{F1}  & \textbf{\#Clstr} \\
\midrule
K-means           &  55.77         & 87.56         & 68.14             & 5k \\
FastHAC           &  99.32         & 64.66         & 78.32             & 117k \\
DBSCAN            &  99.62         & 46.83         & 63.71             & 352k \\
GCN               &  95.87         & 79.43         & 86.88             & 45k \\
GCN-iter2         & 97.94          & 87.28         & 92.30             & 32k \\
\bottomrule
\end{tabular}%
}
\end{table}

\subsection{Controlled Disjoint: MS-Celeb-1M splits} 
\label{sec:ms1m-control}
In controlled setting, Split-1 of MS-Celeb-1M is used as the labeled dataset to train the face recognition model in a fully supervised fashion. The face clustering module is also trained in a supervised way on the labeled Split-1 data. The unlabeled data is from Split-2 of MS-Celeb-1M: ground truth labels are ignored, features are extracted on all the samples and the trained GCN model provides the cluster assignments.

\textbf{Clustering.} The performance of various clustering methods are summarized in Table~\ref{tab:cluster_ms1m}, i.e., \textit{\textbf{K-means}}~\cite{sculley2010web}, \textbf{\textit{FastHAC}}~\cite{mullner2013fastcluster} and \textbf{\textit{DBSCAN}}~\cite{ester1996density,schubert2017dbscan}, with optimal hyper-parameter settings~\footnote{K-means: K=5k, FastHAC: dist=0.85, DBSCAN: minsize=2, eps=0.8}. The GCN is clearly better than the baseline clustering approaches.
GCN typically provides an over-clustering of the actual number of identities -- the precision is comparably higher than the recall (95.87\% versus 79.43\%), indicating high purity per cluster, but samples from the same identity end up being spread out across multiple clusters (``split ID'').


\textbf{Re-training.} The results are summarized in Table~\ref{tab:retrain_ms1m}. Re-training CosFace on labeled Split-1 and pseudo-labeled Split-2 data (\textit{+GCN}) improves over training on just the labeled Split-1 (\textit{Baseline GT-1}) across the benchmarks. The performance is upper-bounded when perfect labels are available on Split-2 (\textit{+GT-2}).
Note that re-training on cluster assignments from simpler methods like K-Means and HAC also improve over the baseline.

\textbf{Re-train w/ iterative clustering.}
We perform a second iteration of clustering, using the re-trained CosFace model as feature extractor. The re-trained CosFace model has more discriminative features, resulting in better clustering (Table~\ref{tab:cluster_ms1m} \textit{GCN-iter2} versus \textit{GCN}). However, another round of re-training CosFace on these cluster-assignments yields smaller gains (Table~\ref{tab:retrain_ms1m} \textit{+GCN-iter2} v.s. \textit{+GCN}).

\textbf{Insights.} With limited labeled data, training on clustered faces significantly improves recognition performance. Simpler clustering methods like K-means are also shown to improve recognition performance -- if training Face-GCN is not practical, off-the-shelf clustering algorithms can also provide pseudo-labels. A second iteration gives small gains, indicating diminishing returns.

\begin{table}[t]
\renewcommand{\tabcolsep}{5pt}
\centering
\small 
\caption{ \small{
\textbf{Controlled disjoint}: Re-training CosFace on the union of labeled and pseudo-labeled data (\textit{+GCN}), pseudo-label on second iteration (\textit{+GCN-iter-2}), with an {\color{CadetBlue}upper bound} from ground truth (\textit{GT-2}). $\uparrow$ indicates improvement from baseline.}
}
\label{tab:retrain_ms1m}
\scalebox{0.9}{
\begin{tabular}{@{\hskip 1mm}l@{\hskip 1mm}c@{\hskip 1mm}c@{\hskip 1mm}c@{\hskip 1mm}c@{\hskip 1mm}c@{\hskip 1mm}c@{\hskip 1mm}c@{\hskip 1mm}c@{\hskip 1mm}}
\toprule
\textbf{Model}     & \textbf{LFW} & $\uparrow$ & \textbf{CFP-fp} & $\uparrow$ & \textbf{IJBA-idt.}  & $\uparrow$  & \textbf{IJBA-vrf.} & $\uparrow$ \\
 & & & & & Rank-1, 5 & & FAR@1e-3,-4 & \\
\midrule
Baseline GT-1   & 99.20 & - & 92.37 & - & 92.66, 96.42 & -  & 80.23, 69.64  & - \\
+K-means       & 99.47 & 0.27 & 94.11 &1.74 & 93.80, 96.79 & 1.14, 0.37 & 87.03, 78.00 & 6.80, 8.36\\
+FastHAC       & 99.42 & 0.22 &  93.56 & 0.90 & 93.84, \textbf{96.81} & 1.18, \textbf{0.39}  & 84.78, 75.21 & 4.55, 5.57 \\
+GCN         & 99.48  & 0.28 & \textbf{95.51} & \textbf{3.14} & 94.11, 96.55 & 1.45, 0.13 & 87.60, 77.67  & 7.37, 7.93 \\
+GCN-\textit{iter-2}  & \textbf{99.57} & \textbf{0.37} & 94.14 &1.77 & \textbf{94.46}, 96.40 & \textbf{1.80}, -0.02 & \textbf{88.00}, \textbf{78.78} & \textbf{7.77}, \textbf{9.14} \\
\midrule
+GT-2 (bound)         & {\color{CadetBlue}99.58} & {\color{CadetBlue}0.38} & {\color{CadetBlue}95.56} & {\color{CadetBlue}3.19} & {\color{CadetBlue}95.24, 97.24} & {\color{CadetBlue}2.58, 0.82} & {\color{CadetBlue}89.45, 81.02} & {\color{CadetBlue}9.22, 11.38}  \\
\bottomrule
\end{tabular}}
\vspace{-4mm}
\end{table}

\begin{table}[tp]
\renewcommand{\tabcolsep}{5pt}
\centering
\small 
\caption{\small{ \textbf{Controlled overlaps:} Re-training with overlapping identity unlabeled data.}}
\label{tab:retrain_ms1m_overlap}
\scalebox{0.9}{
\begin{tabular}{@{\hskip 1mm}l@{\hskip 1mm}c@{\hskip 1mm}c@{\hskip 1mm}c@{\hskip 1mm}c@{\hskip 1mm}c@{\hskip 1mm}c@{\hskip 1mm}c@{\hskip 1mm}c}
\toprule
\textbf{Model}     & \textbf{LFW} & $\uparrow$ & \textbf{CFP-fp} & $\uparrow$  & \textbf{IJBA-idt.}  & $\uparrow$ & \textbf{IJBA-vrf.} & $\uparrow$ \\
 & & & & & Rank-1, 5 & & FAR@1e-3,-4 & \\
\midrule
Baseline &  99.45 & - & 95.17 & - & 94.52, 96.60 & - & 87.36, 75.06 & - \\
+GCN(naive) &  99.37 & -0.08 & 93.17 & -2.0 & 93.72, 96.65 & -0.80, 0.05 & 87.02, 79.39 & -0.34, 4.33 \\
+GCN(disjoint) & 99.57 & 0.12 & 95.01 & -0.16 & \textbf{94.83}, 96.98 & \textbf{0.31}, 0.38 & 89.29, 82.64 & 1.93, 7.58 \\
+GCN(overlap)  & \textbf{99.58} & \textbf{0.13} & 94.30 & -0.87 & 94.47, 96.64 & -0.05, 0.04 & 86.93, 78.42 & -0.43, 3.36 \\
+GCN(both)     & \textbf{99.58} & \textbf{0.13} & \textbf{95.36} & \textbf{0.19} & 94.81, \textbf{97.05} & 0.29, \textbf{0.45} & \textbf{89.43}, \textbf{82.86} & \textbf{2.07}, \textbf{7.80} \\ 
\bottomrule
\end{tabular}}
\end{table}


\begin{table}[t]
\parbox{.45\linewidth}{
\centering
\small 
\renewcommand{\tabcolsep}{5pt}
\caption{\small{
\textbf{Disjoint/overlap clustering.} 
Results of clustering the entire unlabeled $\mathcal{X^{U}}$ (``Split-2-O'') and clustering the estimated ID-disjoint portion $\mathcal{X^{UD}}$.}}
\label{tab:disjoint_cluster}
\scalebox{0.9}{
\begin{tabular}{l@{\hskip 1mm}c@{\hskip 1mm}c@{\hskip 1mm}c@{\hskip 1mm}c@{\hskip 1mm}c@{\hskip 1mm}c}
\toprule
  {Data} &  {Prec.} & {Rec.} & {F1} & {\#IDs} & {\#Clstr} & {\#Img} \\
\midrule 
$\mathcal{X^{U}}$ & 98.7 & 84.8 & 91.2 & 16.8k & 60k & 693k  \\
$\mathcal{X^{UD}}$ & 98.8 & 85.2 & 91.5 & 11.7k & 39k & 464k    \\
\bottomrule
\end{tabular}}%
}
\hfill
\parbox{.5\linewidth}{
\centering
\small 
\caption{\small{\textbf{Semi-controlled: clustering.} 
Comparing performance on ``within-domain'' splits of MS-Celeb-1M vs. VGGFace2 data.}}
\label{tab:cluster_vgg_semi}
\begin{tabular}{@{\extracolsep{5pt}}l@{\hskip 0.5mm}l@{\hskip 0.5mm}c@{\hskip 0.5mm}c@{\hskip 0.5mm}c@{\hskip 0.5mm}r}
\toprule
 \textbf{Train} & \textbf{Test}  & \textbf{Prec.}  & \textbf{Rec.}  & \textbf{F1}  & \textbf{\#clstr}   \\
\midrule
split-1         & split-2       & 95.87          &  79.43        & 86.88             & 45k                  \\
split-1         & VGG2          & 97.65          &  59.62        & 74.04             & 614k                 \\
full            & VGG2          & 98.88          &  72.76        & 83.83             & 224k                 \\
\bottomrule
\end{tabular}
}
\vspace{-6mm}
\end{table}

\subsection{Controlled Overlap: Overlapping Identities} 
\label{sec:id-overlap}
We simulate the real-world overlapping-identity scenario mentioned in Sec.~\ref{sec:overlap} to empirically observe its impact on the ``pseudo-labeling by clustering'' pipeline. We create two subsets of MS1M with around 16k identities each, having about 8.5k overlapping identities (suffix ``\textit{O}'' for \textit{overlaps} in Table~\ref{tab:train_data}). The labeled subset $\mathcal{X}^{L}$ contains around 720k samples (Split-1-O). The unlabeled subset, Split-2-O, contains approximately 467k \textit{disjoint-identity} ($\mathcal{X}^{UD}$) and 224k \textit{overlapping-identity} ($\mathcal{X}^{UO}$) samples.

\textbf{Disjoint/Overlap.} 
Modeling the disjoint/overlapping identity separation as an out-of-distribution problem is an effective approach, especially on choosing the max-logit score as the feature for OOD. A simple Otsu's threshold provides acceptably low error rates, i.e., $\textbf{6.2\%}$ false positive rate and $\textbf{0.69\%}$ false negative rate, while using $95\%$ confidence intervals from Weibulls, we achieve much lower error rates of $\textbf{2.3\%}$ FPR and $\textbf{0.50\%}$ FNR. 

\textbf{Clustering.} Table~\ref{tab:disjoint_cluster} shows the results from clustering all the unlabeled data (\textit{Naive}) versus separating out the identity disjoint portion of the unlabeled data and then clustering (\textit{Disjoint}). On both sets of unlabeled samples, the GCN clustering achieves high precision and fairly high recall, indicating that the clusters we use in re-training the face recognition engine are of good quality. 

\textbf{Re-training.} The results are shown in Table~\ref{tab:retrain_ms1m_overlap}. Naively re-training on the additional pseudo-labels clearly hurts performance (\textit{Baseline} v.s. \textit{GCN(naive)}). Adding pseudo-labels from the \textit{disjoint} data improves over the baseline across the benchmarks. Merging the \textit{overlapping} samples with their estimated identities in the labeled data is done based on the softmax outputs of the baseline model, causing improvements in some cases (\eg~LFW and IJBA verif.) but degrading performance in others (\eg~IJBA ident. and YTF). Merging overlapping identities as well as clustering disjoint identities also shows improvements over the baseline across several benchmarks.

\textbf{Insights.} Overlapping identities with the labeled training set clearly has a detrimental effect when retraining and must be accounted for when merging unlabeled data sources -- the choice of modeling max-logit scores for this separation is shown to be simple and effective. Overall, discarding overlapping samples from re-training, and clustering \textit{only} the disjoint samples, appears to be a better strategy. Adding pseudo-labeled data for classes that exist in the labeled set seems to have limited benefits, versus adding more identities. 


\begin{table}[t]
\renewcommand{\tabcolsep}{5pt}
\centering
\small 
\caption{\small{\textbf{Semi-controlled: MS-Celeb-1M split 1 and VGGFace2}. Note that similar volume of pseudo-labeled data from MS-Celeb-1M split 2 (\textit{+MS1M-GCN-2}) gives lower benefits compared to data from VGGFace2 (\textit{+VGG-GCN}) in challenging settings like IJB-A verification at FAR=1e-4, IJB-A identification Rank-1.}}
\label{tab:retrain_ms1m_vgg_semi}
\scalebox{0.84}{
\begin{tabular}{@{\hskip 0.75mm}l@{\hskip 0.75mm}c@{\hskip 0.75mm}c@{\hskip 0.75mm}c@{\hskip 0.75mm}c@{\hskip 0.75mm}c@{\hskip 0.75mm}c@{\hskip 0.75mm}c@{\hskip 0.75mm}c@{\hskip 0.75mm}c}
\toprule
\textbf{Model}     & \textbf{LFW} & $\uparrow$ & \textbf{CFP-fp} & $\uparrow$  & \textbf{IJBA-idt.}  & $\uparrow$ & \textbf{IJBA-vrf.} & $\uparrow$ \\
 & & & & & Rank-1, 5 & & FAR@1e-3,-4 & \\
\midrule
\small{MS1M-GT-1}          & 99.20  & -   & 92.37  & - & 92.66, 96.42 & - & 80.23, 69.64 & - \\
\small{+MS1M-GCN-2}         & 99.48   & 0.28 & \textbf{95.51} & \textbf{3.14} & 94.11, 96.55 & 1.45, 0.13 & 87.60, 77.67 & 7.37, 12.03 \\
\midrule
\small{+VGG-GCN (ours)}   & \textbf{99.55}   & \textbf{0.35} & 94.60 & 2.23 & \textbf{94.72}, \textbf{96.97} & \textbf{2.06}, \textbf{0.55}  & \textbf{88.12}, \textbf{82.48} & \textbf{7.89}, \textbf{12.84}\\
\midrule
\small{+VGG-GT (bound)}  & {\color{CadetBlue}99.70}  & {\color{CadetBlue}0.50}  & {\color{CadetBlue}97.81} & {\color{CadetBlue}5.44} & {\color{CadetBlue}96.93, 98.25} & {\color{CadetBlue}4.27, 1.83}  & {\color{CadetBlue}93.20, 84.67} & {\color{CadetBlue} 12.97, 15.03} \\ 
\bottomrule
\end{tabular}}
\end{table}

\subsection{Semi-controlled: Limited Labeled, Large-scale Unlabeled Data} 
\label{sec:semi}

MS-Celeb-1M Split 1 forms the labeled data, while the unlabeled data is from VGGFace2 (Table~\ref{tab:train_data}). We simply discard VGGFace2 samples estimated to have overlapping identities with MS-Celeb-1M Split-1. Out of the total 3.1M samples, about 2.9M were estimated to be identity-disjoint with MS-Celeb-1M Split-1.

\textbf{Clustering.} 
The same GCN model trained on Split-1 of MS-Celeb-1M in Sec.~\ref{sec:ms1m-control} is used to obtain cluster assignments on VGGFace2. Table~\ref{tab:cluster_vgg_semi} compares the clustering on MS-Celeb-1M Split2 (controlled) v.s. the current setting.
The F-score on VGGFace2 is reasonable -- 74.04\%, but lower than the F-score on Split-2 MS-Celeb-1M (86.88\%) -- we are no longer dealing with within-dataset unlabeled data. 

\textbf{Re-training.} 
To keep similar volumes of labeled and pseudo-labeled data we randomly select 50 images per cluster from the largest 8.5k clusters of VGGFace2.
Re-training results are in Table~\ref{tab:retrain_ms1m_vgg_semi}. We generally see benefits from VGGFace2 data over both \textit{baseline} and \textit{MS1M-split-2}: YTF: 93.82\% $\rightarrow$ 94.64\% $\rightarrow$ \textbf{95.14\%}, IJBA idnt. rank-1: 92.66\% $\rightarrow$ 94.11\% $\rightarrow$ \textbf{94.72\%}, IJBA verif. at FAR 1e-4: 69.635\% $\rightarrow$ 77.665\%  $\rightarrow$ \textbf{82.484\%}. 
When the full VGGFace2 labeled dataset is used to augment MS1M-split-1, \textit{VGG-GT(full)}, we get the upper bound performance.

\textbf{Insights.}
Ensuring the \textit{diversity} of unlabeled data is important, in addition to other concerns like clustering accuracy and data volume: pseudo-labels from VGGFace2 benefit more than using more data from within MS1M.

\subsection{Soft Labels for Clustering Uncertainty} 
\label{sec:soft_label}
Table~\ref{tab:soft_retrain} shows results of re-training the face recognition model with our proposed cluster-uncertainty weighted loss (Sec.~\ref{sec:retrain}) on the pseudo-labeled samples (\textit{GCN-soft}). We set $\gamma = 1$ (ablation in supplemental).
%
We empirically find that incorporating this cluster uncertainty into the training loss improves results in both controlled and large-scale settings (3 out of 4 evaluation protocols).
In the controlled setting, the soft pseudo-labels, MS1M-GCN-\textit{soft}, improves over MS1M-GCN (hard cluster assignments) on challenging IJB-A protocols (77.67\% $\rightarrow$ \textbf{78.78}\% @FAR 1e-4) and is slightly better on LFW. 
In the large-scale setting, comparing VGG-GCN and VGG-GCN-\textit{soft}, we again see significant improvements on IJB-A (81.85\% $\rightarrow$ \textbf{90.16}\% @FAR 1e-4) and gains on the LFW benchmark as well.
%
%
Qualitative analyses of clustering errors and uncertainty estimates $p^-(\mathbf{x}_i)$ are included in the supplemental.

\begin{table}[t]
\renewcommand{\tabcolsep}{5pt}
\centering
\small 
\caption{\small{\textbf{Effect of Cluster Uncertainty}: Re-training CosFace with the proposed clustering uncertainty (\textit{GCN-soft}) shows improvements in both controlled (\textit{MS1M-GT-split1}) and large-scale settings, \textit{MS1M-GT-full} (CosFace~\cite{wang2018cosface}).}}
\label{tab:soft_retrain}
\scalebox{0.82}{
\begin{tabular}{@{\hskip 0.75mm}l@{\hskip 0.75mm}c@{\hskip 0.75mm}c@{\hskip 0.75mm}c@{\hskip 0.75mm}c@{\hskip 0.75mm}c@{\hskip 0.75mm}c@{\hskip 0.75mm}c@{\hskip 0.75mm}c@{\hskip 0.75mm}c}
\toprule
\textbf{Model}     & \textbf{LFW} & $\uparrow$  & \textbf{CFP-fp} & $\uparrow$ & \textbf{IJBA-idt.} & $\uparrow$   & \textbf{IJBA-vrf.} & $\uparrow$ \\
 & & & & & Rank-1, 5 & & FAR@1e-3,-4 & \\
\midrule
\small{MS1M-GT-\textbf{\textit{split1}}}   & 99.20 & - & 92.37 & - & 92.66, 96.42 & - & 80.23, 69.64 & - \\
\small{+MS1M-GCN (ours)} & 99.48 & 0.28 & \textbf{95.51} & \textbf{3.14} & 94.11, 96.55 & 1.45, 0.13 & 87.60, 77.67 & 7.37, 12.03  \\
\small{+MS1M-GCN-\textit{soft} (ours)}  & \textbf{99.50} & \textbf{0.30} & 94.71 & 2.34 & \textbf{94.76}, \textbf{97.10} & \textbf{2.10}, \textbf{0.68} & \textbf{87.97}, \textbf{79.43} & \textbf{7.74}, \textbf{9.79} \\
\midrule
\small{MS1M-GT-\textbf{\textit{full}} (CosFace)} & 99.70 & - & \textbf{98.10} & - & 95.47, 97.04 & - & 92.82, 80.68 & - \\
\small{+VGG-GCN (ours)} & 99.73 & 0.03 & 97.63 & -0.47 & 95.87, 97.45 & 0.40, 0.41  & 93.88, 81.85 & 1.06, 1.17 \\
\small{+VGG-GCN-\textit{soft} (ours)} & \textbf{99.75} & \textbf{0.05} & 97.57 & -0.53 & \textbf{96.37}, \textbf{97.70} & \textbf{0.90}, \textbf{0.66} & \textbf{93.94}, \textbf{90.16} & \textbf{1.12}, \textbf{9.48} \\
\bottomrule
\end{tabular}}
\end{table}

\begin{table}[t]
\renewcommand{\tabcolsep}{5pt}
\centering
\small 
\caption{\small{\textbf{Uncontrolled: pseudo-labels.} Showing the clusters and samples in the uncontrolled setting with full-MS1M and unlabeled data of increasingly larger volume -- (1) VGG2~\cite{cao2018vggface2}; (2) merging CASIA~\cite{yi2014learning} \& IMDB-SenseTime~\cite{wang2018devil} with VGG2; (3) merging GlintAsian~\cite{glintasian} with all the above. 
}}
\label{tab:uncontrol_cluster}
\begin{tabular}{@{\extracolsep{5pt}}lrrr}
\toprule
 \textbf{Dataset:}  & VGG2     & +(CASIA, IMDB) &  +Glint \\
\midrule
  True classes      & 8631        & 57,271    &   149,824\\
  Clusters          & 224,466     & 452,598    &  719,722 \\
  Samples           & 1,257,667   & 2,133,286    &  3,673,517 \\
\midrule 
  Prec.             & 98.88       & 91.35    &  88.16 \\
  Rec.              & 72.76       & 77.53    &  66.93 \\
  F-score           & 83.83       & 83.88    &  76.09 \\
\bottomrule
\end{tabular}
\end{table}


\begin{table}[t]
\renewcommand{\tabcolsep}{5pt}
\centering
\small 
\caption{\small{\textbf{Uncontrolled: re-training.} Merging unlabeled training samples with the entire MS-Celeb-1M labeled data consistently surpasses the fully-supervised MS1M-GT-\textit{full} (CosFace~\cite{wang2018cosface}) trained on the entire labeled MS-Celeb-1M dataset.}}
\label{tab:full_ms1m_vgg-gcn}
\scalebox{0.80}{
\begin{tabular}{@{\hskip 0.75mm}l@{\hskip 0.75mm}c@{\hskip 0.75mm}c@{\hskip 0.75mm}c@{\hskip 0.75mm}c@{\hskip 0.75mm}c@{\hskip 0.75mm}c@{\hskip 0.75mm}c@{\hskip 0.75mm}c@{\hskip 0.75mm}c}
\toprule
\textbf{Model}     & \textbf{LFW} & $\uparrow$ & \textbf{CFP-fp} & $\uparrow$ & \textbf{IJBA-idt.} & $\uparrow$ & \textbf{IJBA-vrf.} & $\uparrow$  \\
 & & & & & Rank-1, 5 & & FAR@1e-3,-4 & \\
\midrule
\small{MS1M-GT-\textbf{\textit{full}} (CosFace)}     & 99.70 & - & 98.10 & - & 95.47, 97.04 & - & 92.82, 80.68 & - \\
\small{+VGG-GCN (ours)} & \textbf{99.73} & \textbf{0.03} & 97.63 & -0.47 & 95.87, 97.45 & 0.40, 0.41 & 93.88, 81.85 & 1.06, 1.17 \\
\small{+CASIA-IMDB (ours)} & \textbf{99.73} & \textbf{0.03} & 97.81 & -0.29 & 96.66, 97.89 & 1.19, 0.85 & 93.79, 89.58 & 0.97, 8.90 \\
\small{+GlintAsian (ours final)} & \textbf{99.73} & \textbf{0.03} &  \textbf{98.24} & \textbf{0.14} & \textbf{96.94}, \textbf{98.21} & \textbf{1.47}, \textbf{1.17} & \textbf{94.89}, \textbf{92.29} & \textbf{2.07}, \textbf{11.61} \\
\bottomrule
\end{tabular}}
\end{table}



\subsection{Uncontrolled: Large-scale Labeled and Unlabeled Data} 
\label{sec:uncontrol_vgg}

The earlier cases either had limited labeled data, unlabeled data from an identical distribution as the labeled data by construction, or both aspects together. Now, the \textit{entire} MS-Celeb-1M is used as labeled training data for training the baseline CosFace model as well as the GCN.
We gradually add several well-known face recognition datasets (ignoring their labels) to MS-Celeb-1M labeled samples during re-training (Table~\ref{tab:uncontrol_cluster})~\footnote{In particular, we estimated a 40\% overlap in identities between MS-Celeb and VGG2.}. 
Along with more data, these datasets bring in more \textit{varied} or
\textit{diverse} samples (analysis in supplemental).

\textbf{Re-training.}
The re-training results are shown in Table~\ref{tab:full_ms1m_vgg-gcn}. As expected, we get limited benefits from adding moderate amounts of unlabeled data when the baseline model is trained on a large labeled dataset like MS-Celeb-1M. When incorporating data from only VGGFace2, there are improvements on LFW (99.7\% $\rightarrow$ 99.73\%), and on IJBA, ident. (95.47\% $\rightarrow$ 95.87\%) and verif. (80.68\% $\rightarrow$ 81.85\%). There are however some instances of decreased performance on the smaller scale dataset CFP-fp.
When the volume of unlabeled data is of comparable magnitude (4.7M labeled versus 3.6M unlabeled) by merging all the other datasets (VGGFace2, CASIA, IMDB-SenseTime and GlintAsian), we get a clear advantage on the challenging IJBA benchmarks (rank-1 identification: 95.47\% $\rightarrow$ \textbf{96.94\%}, verification TAR at FAR 1e-4: 80.68\% $\rightarrow$ \textbf{92.29\%}). 

\textbf{Insights.}
The crucial factors in improving face recognition when we have access to all available labeled data from MS1M appear to be \textit{both} diversity and volume -- it is only when we merged unlabeled data from all the other data sources, reaching comparable number of samples to MS1M, that we could improve over the performance attained from training on just the ground-truth labels of MS1M, 
%
suggesting that current high-performing face recognition models can benefit from even larger training datasets.
While acquiring datasets of such scale purely through manual annotation is prohibitively expensive and labor-intensive, using pseudo-labels is shown to be a feasible alternative.



\vspace{-2.5mm}
\section{Conclusion}
\label{sec:conclusion}
\vspace{-4mm}
\noindent
The pseudo-labeling approach described in this paper provides a recipe for improving fully supervised face recognition, i.e., CosFace, leveraging large unlabeled sources of data to augment an existing labeled dataset. The experimental results show consistent performance gains across various scenarios and provide insights into the practice of large-scale face recognition with unlabeled data -- \textbf{(1)} we require comparable volumes of labeled and unlabeled data to see significant performance gains, especially when several million labeled samples are available; \textbf{(2)} overlapped identities between labeled and unlabeled data is a major concern and needs to be handled in real-world scenarios; \textbf{(3)} along with large amounts of unlabeled data, greater gains are observed if the new data shows certain domain gap w.r.t. the labeled set; \textbf{(4)} incorporating scalable measures of clustering uncertainty on the pseudo-labels is helpful in dealing with label noise.
Overall, learning from unlabeled faces is shown to be an effective approach to further improve face recognition performance.

\noindent {\bf Acknowledgments.} 
This research was partly sponsored by the AFRL and DARPA under agreement FA8750-18-2-0126. The U.S.~Government is authorized to reproduce and distribute reprints for Governmental purposes notwithstanding any copyright notation thereon. The views and conclusions contained herein are those of the authors and should not be interpreted as necessarily representing the official policies or endorsements, either expressed or implied, of the AFRL and DARPA or the U.S.~Government.


{\footnotesize
\bibliographystyle{splncs04}
\bibliography{egbib_short}
}


\end{document}




\pagestyle{headings}
\mainmatter
\def\ECCVSubNumber{4531}

\title{
Supplementary Materials:\\
Improving Recognition with\\
Unlabeled Faces in the Wild
}

\titlerunning{Improving Face Recognition by Clustering Unlabeled Faces in the Wild}
%
\author{Aruni RoyChowdhury\inst{1} \and
Xiang Yu\inst{2} \and
Kihyuk Sohn\inst{2} \and \\
Erik Learned-Miller\inst{1} \and
Manmohan Chandraker \inst{2}
}
\authorrunning{A. RoyChowdhury et al.}
%
\institute{University of Massachusetts Amherst \and
NEC Labs America
}

\maketitle
\thispagestyle{empty}

We include some additional experimental details and discussions here that could not be included in the main paper due to space constraints:
\begin{itemize}
        \item{The overview of approaches on improving face recognition (Sec.~\ref{sec:overview})}
        
        \item{The diversity in datasets (Sec.~\ref{sec:frechet})}
        
        \item{Discussion on the motivation of our choice to model the label noise (Sec.~\ref{sec:uncertain})} 
        
        \item{The effect of the $\gamma$ hyper-parameter on our proposed uncertainty weighted loss  (Sec.~\ref{sec:gamma})} 
        
        \item{Details and baselines for overlapping identity separation (Sec.~\ref{sec:overlap_gaus})}
        
        \item{Visualization of clustering errors and their correspondence to uncertainty scores (Sec.~\ref{sec:cluster_viz})}
        
        \item{Detailed descriptions of the evaluation benchmarks (Sec.~\ref{sec:eval})} 
        
        \item{Implementation details including training settings for the clustering module and the deep face networks (Sec.~\ref{sec:implement})} 
        
\end{itemize}

\section{Overview}
\label{sec:overview}
The main submission empirically illustrated the use of unlabeled faces to improve fully-supervised face recognition systems. From the literature, one major direction to boost performance is via \textit{supervised} training, \ie,  leverage various network structures such as VGG Face~\cite{simonyan14very}, ResNet~\cite{he2016deep} and SE-Net~\cite{hu2018squeeze}, or investigating effective objective functions, \ie, triplet loss~\cite{schroff2015facenet}, Cosine Loss~\cite{wang2018cosface}, by constraining the feature lying on a hypersphere~\cite{liu2017sphereface}, or further combine the two~\cite{deng2018arcface}.

Our paper advocates another direction: leverage larger amounts of unlabeled training data in a \textit{semi-supervised} manner. These two axes lead to \textit{orthogonal} developments -- more data is likely to improve the next generation of better face architectures and losses. Moreover, tasks such as automatic adaptation of a model to a new scene or condition will benefit from being able to learn from unlabeled faces. There are several use cases for such adaptation: \eg, a particular ethnicity may not have a large labeled dataset but have many unlabeled faces available. In general, deployed models would be able to leverage a continuous stream of unlabeled data to adapt to specific operational conditions.

We briefly re-iterate our main conclusions here -- the experiments show that it is indeed possible to further improve the recognition performance of fully-supervised models by exploiting clustering to obtain pseudo-labeled additional data. To see significant improvements, we require comparable volumes of labeled and pseudo-labeled data, as well as accounting for label noise and overlapped identities between labeled and unlabeled sets.

\section{On quantifying the diversity in data}
\label{sec:frechet}

\begin{figure}[!t]
    \centering
    \includegraphics[width=0.6\textwidth]{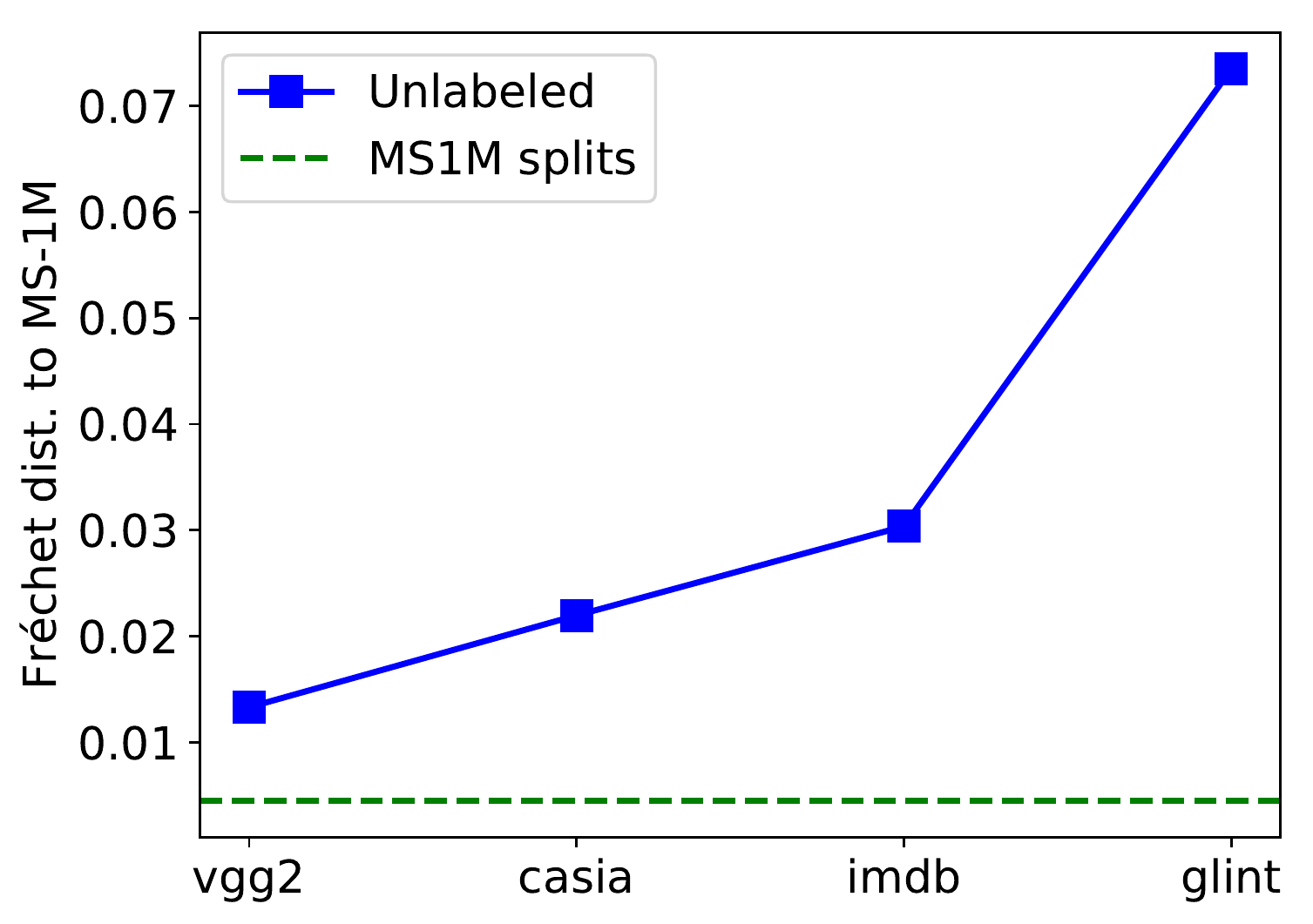}
    \caption{\small{Fr\'echet distance~\cite{dowson1982frechet,heusel2017gans} between MS-1M and the other datasets used as sources of unlabeled data, indicating the domain gap between MS-1M and the unlabeled data.
    }}
    \label{fig:frechet}
\end{figure}

In the large-scale uncontrolled setting experiments presented in the main paper, we observe that samples from a different dataset provides greater benefits w.r.t. performance than more samples drawn from the same distribution as the original labeled training dataset. Intuitively this makes sense -- different datasets would bring in more information that the original network has not seen earlier during training. We quantify this notion of distance or diversity among datasets using a simple Fr\'echet distance~\cite{dowson1982frechet,heusel2017gans}, visualized in  Fig.~\ref{fig:frechet}. The different datasets are arranged along the x-axis based on distance from labeled MS-Celeb.

\section{Motivation of Modeling the Label Noise}
\label{sec:uncertain}

Here, we begin by discussing some related work regarding our choice of modeling the noisy labels that may motivate our choice of using a \textit{linear} classifier on top of face descriptor features to decide which cluster assignments are noisy. We note that the label noise from clustering is well-structured, very much unlike the uniform noise (\ie all categories are equally likely to have their correct label flipped) well-studied in the literature on neural network generalization~\cite{zhang2016understanding,arpit2017closer,Forgetting}. 
Zhang~\etal~\cite{zhang2016understanding} show that deep neural networks are able to perfectly memorize random labels assigned to the training samples. This would indicate that a network of sufficient expressivity would be able to memorize the incorrect labels in our pseudo-labeled dataset, leading to sub-optimal performance upon re-training with the extra data. Arpit~\etal~\cite{arpit2017closer} however observe that despite the ability to memorize random patterns, deep neural networks tend to learn easy or correctly-labeled patterns first, and then start fitting to the incorrectly labeled examples in subsequent training epochs. \cite{ICML2019_UnsupervisedLabelNoise} report that the training loss of a network on noisy labeled samples is higher than correctly labeled training samples, and this difference can be used to separate out the noisy labels. 

We observe in our initial experiments that at least on our face datasets, the highly-structured labeled noise from clustering assignments behaves differently -- even shallower neural networks were learning to fit to both incorrectly and correctly labeled samples at almost concurrent rates, and thus there was no clear separation by looking at the empirical distribution of the training loss. 
\textit{Mixup}~\cite{zhang2017mixup} shows that encouraging deep neural networks to behaving linearly in between samples improves generalization and tolerance to noise.
In fact, \cite{ICML2019_UnsupervisedLabelNoise} report \textit{mixup} regularization to be useful in their label noise robustness experiments. 

\begin{table}[t]
\renewcommand{\tabcolsep}{5pt}
\centering
\small 
\caption{Effect of tuning hyper-parameter $\gamma$ on the uncertainty weighted loss.
}
\label{tab:gamma_ms1m}
\begin{tabular}{@{\hskip 2mm}l@{\hskip 2mm}c@{\hskip 2mm}c@{\hskip 2mm}c@{\hskip 2mm}c@{\hskip 2mm}c}
\toprule
\textbf{Model}     & \textbf{LFW}  & \textbf{CFP-fp} & \textbf{IJBA-idt.}    & \textbf{IJBA-vrf.}  \\
 & & & Rank-1, 5 & FAR@1e-3,-4 \\
\midrule
Baseline GT-1   & 99.20 & 92.37 & 92.66, 96.42  & 80.23, 69.64  \\
+ GCN           & 99.48 & {95.51} & 94.11, 96.55 & 87.60, 77.67  \\
\midrule
+ GCN $\gamma = \frac{1}{3}$  & {99.60}  & {94.66}    & 94.73, 96.93     &  87.93, 81.16  \\
+ GCN $\gamma = \frac{1}{2}$  & {99.45}  & {92.86}    & 93.47, 96.44     &  84.13, 75.26 \\
+ GCN $\gamma = 1$            & {99.50} & {94.71}     & 94.76, 97.10     &  87.97, 79.43  \\
+ GCN $\gamma = 2$            &  99.48  & {94.71}     & 95.05, 97.26     &  88.43, 79.87  \\
+ GCN $\gamma = 3$            &  99.55  &  94.47      & 94.88, 97.24     &  88.12, 78.74  \\
\midrule
+ GT-2 (bound)    & 99.58 & 95.56 & 95.24, 97.24 & 89.45, 81.02  \\
\bottomrule
\end{tabular}
\end{table}

Our intuition for using linear separability to estimate label noise is as follows -- assuming that effective features have been learned by the baseline model on a large labeled dataset, we trust \textit{only} those cluster assignments that can be fitted by a simple linear classifier on top of these discriminative features. While this does reduce the opportunity of the deep network to learn from some challenging examples (\ie complicated clusters which are not modeled by a simple linear model would have a high loss that may benefit the network), it also reduces the chance of the high losses from incorrectly-clustered samples from destabilizing the network training. 

\section{Effect of Hyper-parameter $\gamma$}
\label{sec:gamma}
Setting various values of $\gamma$ in the weighted loss can change the steepness of the weighting curve following a power law: 
$$\mathcal{L}^{p}(\mathbf{x}_i) = (1-p^-(\mathbf{x}_i))^{\gamma}\mathcal{L}(\mathbf{x}_i)$$

The behaviour is somewhat like the ``focusing parameter'' in methods like the focal-loss~\cite{lin2017focal}. However, despite some similarities, the motivation and the implementations are starkly different -- focal loss seeks to emphasize high-loss samples in a training batch, as a means of hard-example mining; we seek to discount the effect of samples which we suspect are incorrectly pseudo-labeled. Moreover, the focal loss uses the deep network's softmax output as the posteriors, while we have a separate parametric model to estimate the probability of an incorrect label.
We show the re-training performance at different values of $\gamma$ in the uncertainty-weighted loss in Table~\ref{tab:gamma_ms1m}. The parametric Weibull model on the classification-margin appears to be a good estimate of this uncertainty, and changing the shape of the curve gives limited benefits. The focusing parameter is observed to have limited effect in practice -- the improvements are not consistent across datasets, and therefore we simply use $\gamma = 1$ in all further experiments. We note that other choices than Weibull, \eg Laplace or beta~\cite{ICML2019_UnsupervisedLabelNoise}, may be used to parameterize this distribution -- our choice was based on the observed skewness of the empirical distribution, which precluded the more common Gaussian.

\section{Overlapping Identity Separation}
\label{sec:overlap_gaus}


We show the results of modeling the disjoint/overlapping identity separation as an out-of-distribution problem in Table~\ref{tab:disjoint_overlap}.
These results were presented in a much condensed form in the main paper.
A simple Otsu's threshold provides acceptably low error rates, i.e., $6.2\%$ false positive rate and $0.69\%$ false negative rate. This shows that our choice of the max-logit score as the feature for OOD is an effective approach.

\begin{table}[t]
\renewcommand{\tabcolsep}{5pt}
\centering
\small 
\caption{\textbf{Separating overlapped identities.} Results on detecting samples in the unlabeled data whose identity overlaps with classes in the labeled training set.
}
\label{tab:disjoint_overlap}
\begin{tabular}{@{\extracolsep{4pt}}lccc}
\toprule
 \textbf{Method}    & \textbf{False Positives}  & \textbf{False Negatives} & \textbf{SSE} \\
\midrule 
  Naive Otsu        & 6.2\%     & 0.69\%    & - \\
  Gaussian-95\%     & 2.01\%    &   0.51\%  & 0.245 \\
  Weibull-95\%      & 2.33\%    & 0.50\%    & 0.228 \\
\bottomrule
\end{tabular}
\end{table}

\begin{figure}[t]
    \centering
      \includegraphics[width=0.8\textwidth]{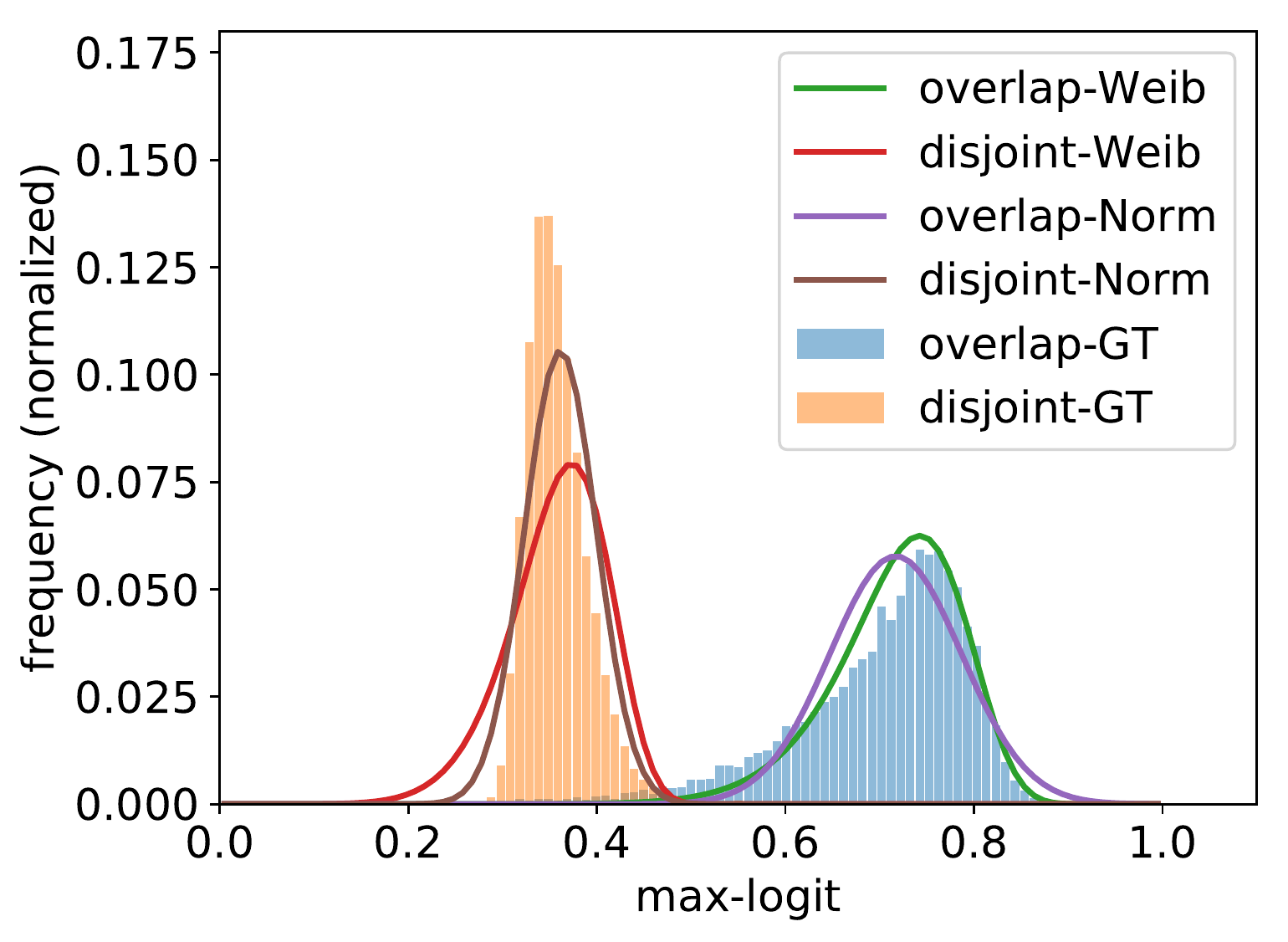}
    \caption{Empirical distribution of the max-logit score for overlapping and disjoint identities between labeled and unlabeled sets (shown on controlled splits of MS-1M dataset). The two-component Weibull and Gaussian models are shown in solid lines.}
    \label{fig:overlap_gaussian}
\end{figure}

Fig.~\ref{fig:overlap_gaussian} shows the Weibull and a baseline Gaussian model fit to the empirical distributions of max-logit scores. We quantify the error in fitting the actual data by the sum-of-squared-errors (SSE) between empirical and theoretical PDFs, shown in the last column of Table~\ref{tab:disjoint_overlap}. The Gaussian model has a slighly higher SSE, indicating a worse fit overall. This justifies the decision to fit the maxima using the Weibull family.

Using $95\%$ confidence intervals from Weibulls, we achieve much lower error rates than the simple Otsu's threshold: ${2.3\%}$ FPR and ${0.50\%}$ FNR. 
Using Gaussians to threshold the max-logits gives almost equivalent results for overlap separation (slightly better in FP and worse in FN), although the Weibulls fit the skewed distributions better.

\section{Visualization of Clustering Errors} 
\label{sec:cluster_viz}

As discussed in the main paper, our goal is to obtain a model of the noise that can capture the structured label noise resulting from clustering. Re-iterating our steps: \textbf{(1)} train a linear classifier on cluster assignments; \textbf{(2)} define metrics of classification uncertainty such as entropy, classification-margin etc.; \textbf{(3)} to validate the hypothesis, check how well this uncertainty metric corresponds to clustering errors.

\begin{figure}[t]
    \centering
    \begin{tabular}{cc}
      \vspace{-0.1cm}
      \includegraphics[width=0.45\textwidth,trim=0 0 0 1cm, clip]{figures/noisy-label/cluster_noise.pdf} & 
      \includegraphics[width=0.45\textwidth]{figures/noisy-label/pr-curve_log-reg.pdf} \\
      (a) & (b) \\
      \includegraphics[width=0.45\textwidth]{figures/noisy-label/weibull-loss-hist_ms1m-gcn-split-2_err.pdf} &
      \includegraphics[width=0.45\textwidth]{figures/noisy-label/soft-score_ms1m-gcn-split-2_err.pdf} \\
      (c) & (d) \\
    \end{tabular}
    \caption{ \textbf{Clustering uncertainty.}
    \textbf{(a)} Examples of incorrect pseudo-labels -- an image of \textit{George Bush Sr.} is included in a cluster of \textit{George W Bush} images (outlier circled in {\color{blue}blue}); some \textit{George W Bush} images are spread across multiple clusters (``split ID'' circled in {\color{red}red}).
    \textbf{(b)} Precision-recall curves showing Average Precision (AP) of predicting if a cluster assignment is correct using class-margin, max-logit and entropy. 
    \textbf{(c)} Distribution of class-margin with a Weibull fit to the left mode (orange curve).
    \textbf{(d)} An importance weight is assigned to each pseudo-labeled sample based on its likelihood under the Weibull. 
    }
    \label{fig:cluster-err}
\end{figure}

\begin{figure*}[!!tp]
    \centering
     \qquad Inliers \qquad \qquad  \qquad  \qquad  \qquad  \qquad Outliers \\
    \includegraphics[width=0.65\textwidth]{latex/figures/images/cluster-noise-outliers/cluster_noise_faces-outliers_1.pdf} \\
    \includegraphics[width=0.65\textwidth]{latex/figures/images/cluster-noise-outliers/cluster_noise_faces-outliers_2.pdf} \\
    \includegraphics[width=0.65\textwidth]{latex/figures/images/cluster-noise-outliers/cluster_noise_faces-outliers_3.pdf} \\
    \includegraphics[width=0.65\textwidth]{latex/figures/images/cluster-noise-outliers/cluster_noise_faces-outliers_4.pdf} \\
    \includegraphics[width=0.65\textwidth]{latex/figures/images/cluster-noise-outliers/cluster_noise_faces-outliers_5.pdf} \\
    \caption{ \textbf{Cluster outliers.} The \textbf{\textit{left column}} shows inlier samples from 5 clusters. The \textbf{\textit{right column}} shows faces of \textit{different identity} being assigned to the \textit{same cluster} as on the left (outlier samples).  
    The numbers below show the mean and standard deviation of the likelihood of being a noisy label ($p^-$). Note that the outlier samples on the right on average have significantly higher likelihood under this noise model. 
    Having a distinctive common attribute like eye-glasses (\textit{row 2}), facial hair (\textit{row 3}) or prominent eyebrows (\textit{last row}) can confound the clustering, even though the identities are different.
     }
    \label{fig:cluster-outlier}
\end{figure*}

\begin{figure*}[tp]
    \centering
    \qquad True ID \qquad \qquad  \qquad  \qquad  \qquad  \qquad Split-ID \\
    \includegraphics[width=0.65\textwidth]{figures/images/cluster-noise-split-id/cluster_noise_faces-split-ID_1.pdf} \\
    \includegraphics[width=0.65\textwidth]{figures/images/cluster-noise-split-id/cluster_noise_faces-split-ID_2.pdf} \\
    \includegraphics[width=0.65\textwidth]{figures/images/cluster-noise-split-id/cluster_noise_faces-split-ID_3.pdf} \\
    \includegraphics[width=0.65\textwidth]{figures/images/cluster-noise-split-id/cluster_noise_faces-split-ID_4.pdf} \\
    \includegraphics[width=0.65\textwidth]{figures/images/cluster-noise-split-id/cluster_noise_faces-split-ID_5.pdf}
    \caption{ \textbf{Split-identity clustering.} The \textbf{\textit{left column}} shows samples from 5 clusters. The \textbf{\textit{right column}} shows faces that share the \textit{same identity} as on the left, but have been assigned to \textit{different clusters}. The numbers below show the mean and standard deviation of the likelihood of being a noisy label ($p^-$). Note that the ``split identity'' samples on the right, that have been separated from the ``true cluster'' of that identity, have higher values under this noise model.
    E.g. \textit{top row:} all images belong to the same person (actor Max von Sydow), but due to factors such as age and facial hair, all images are not assigned the same cluster.
     }
    \label{fig:cluster-split-id}
\end{figure*}

To this end, we attempt to quantify the typical errors that occur in cluster assignments (Fig.~\ref{fig:cluster-err})~\footnote{We repeat this figure here from the main paper for ease of exposition in the writing.}, based off the standard metrics of precision and recall:

\begin{itemize}
        \item{\textbf{\textit{Outliers:}} Using ground-truth labels, we first find the modal or most frequent  identity in a cluster. Samples corresponding to this identity are \textit{inliers}. The others are \textit{outliers}. This type of error affects the precision of the clustering algorithm. Some illustrative examples from the MS-1M splits are shown in Fig.~\ref{fig:cluster-outlier}, where each row depicts a cluster. The clustering algorithm confuses matching attributes like facial hair, sunglasses, heavy eyebrows etc. for identity, and ends up putting different people into the same cluster.}
        
        \item{\textbf{\textit{Split-identity:}} This type of error occurs when samples from the \textit{same} identity as split across \textit{different} clusters, which impacts the recall metric of a clustering algorithm. For a ground-truth identity, we find all clusters that contain samples belonging to this identity. A perfect clustering would assign all samples of a person to a \textit{single} cluster, but this is generally not the case -- samples of a person can be scattered or split over several clusters~\footnote{Note that Face-GCN typically has very high precision, but comparatively lower recall, which is why this type of error is more common in our experiments.}. We find the cluster with the highest number of samples for a particular identity, regarding it as the ``true cluster'', and the other clusters as having incorrectly split the identity (this is a rough heuristic that we empirically found to be feasible). Some examples of this scenario are shown in Fig.~\ref{fig:cluster-split-id}. E.g. the first row shows various images of the Swedish actor \href{https://www.imdb.com/name/nm0001884/?ref_=nmbio_bio_nm}{Max von Sydow}. Most of his middle-aged and older images form the largest or ``true'' cluster, shown on the left. Several images that exhibit other attributes like facial hair or a much younger age end up forming separate clusters, as shown on the right.} 
\end{itemize}

As detailed in the main paper, we use precision-recall curves to analyse the correspondence of our uncertainty metrics with clustering errors, finding the highest Average Precision (AP) with \textit{\textbf{classification-margin}} (95.16\%), with \textit{max-logits} and softmax \textit{entropy} getting APs of 94.80\% and 88.29\%, respectively. The empirical distribution of classification-margin scores on a noisy dataset was observed to be bi-modal -- incorrect clusterings had a small classification-margin since they were difficult for the logistic regression classifier to learn correctly. In Fig.~\ref{fig:cluster-err}(b), a Weibull distribution fit to the lower mode gives our noise model $p^{-}(\mathbf{x}_i)$, \ie the likelihood that a sample $\mathbf{x}_i$ has been clustered incorrectly. Figures~\ref{fig:cluster-outlier} and \ref{fig:cluster-split-id} also show the average values of $p^{-}(\mathbf{x}_i)$ for the samples -- inliers and true-clusters are typically given a lower likelihood under this model, \ie we are \textit{less} uncertain about their cluster assignment.

\section{Evaluation Benchmarks}
\label{sec:eval}
The main paper presents results on the following benchmarks, which we describe in more details here: 

\begin{itemize}
    \item{\textit{\textbf{Labeled Faces in the Wild (LFW)}}~\cite{huang2008labeled,learned2016labeled}: consists of 13,233 images and 5749 people, reporting verification accuracy across 10 folds of 300 matching and 300 non-matching face pairs.} 
    
    \item{\textit{\textbf{Celebrity Frontal to Profile (CFP)}}~\cite{sengupta2016frontal}: consists of 500 people, each with 10 frontal and 4 profile images. There are two verification protocols -- frontal to frontal (\textit{ff}) and frontal to profile (\textit{fp}) images. Each protocol consists of 10 folds with 350 same-identity and mismatched-identity pairs.}
    
    
    \item{\textit{\textbf{IJB-A}}~\cite{klare2015pushing}:  part of the challenging IARPA Janus benchmark, it has 500 subjects with 5,397  images and 2,042 videos. Identification performance is reported as retrieval rate at ranks 1 and 5, using 10 splits each with 112 gallery templates and 1763 probe templates (\ie 1,187  genuine queries and 576 impostor queries whose identities are not in the gallery). Verification performance is reported as True Accept Rate (TAR) at False Accept Rates (FAR) ranging from 1e-1 to 1e-4, evaluated on 10 splits with 11,748 pairs of templates (1,756 positive and 9,992 negative pairs); we report performance at the two most strict settings: FAR@1e-3,1e-4 respectively.} 
\end{itemize}

\section{Implementation Details}
\label{sec:implement}
\noindent
\textbf{Face recognition training.} 
The CosFace model~\cite{wang2018cosface} is used as our face recognition engine, which is one of the top performance methods on standard face recognition benchmarks. A 118-layer ResNet is used as the backbone network. 
The baseline model on labeled data is trained for 30 epochs using SGD with momentum $0.95$, with a batch size of 512 across 8 GPUs in parallel, starting from a learning rate of 0.1, with the learning rate dropping by a factor of $1/10$ at the $16^{th}$ and $23^{rd}$ epochs.  When used as a feature extractor, this model yields vectors of 512 dimensions. When training with pseudo-labeled data, we re-train the entire model from scratch on the union of the labeled and pseudo-labeled data, with the same training settings. 

\noindent
\textbf{Clustering model training.} The Face-GCN implementation uses the publicly available code~\footnote{\url{https://github.com/yl-1993/learn-to-cluster}} of GCN-D from \cite{yang2019learning}. An initial k-nearest neighbor graph is formed over the unlabeled samples with $k=80$, using the \href{https://github.com/facebookresearch/faiss/wiki}{FAISS} library for efficient similarity computation over large sample sizes. Cluster proposals are generated from this by setting various thresholds -- we find optimal settings on a held-out set of MS-Celeb-1M and continue to use these consistently on all the other datasets. The \texttt{GCN-D} model from Face-GCN is trained to predict the precision and/or recall for each cluster proposal. We use a simple 3-layer architecture, with feature sizes: $512 \rightarrow 256 \rightarrow 64$, following by a global max-pooling. Following \cite{yang2019learning}, the model is trained with a regression loss.

\noindent
\textbf{Re-training on pseudo-labels.} Following the final clustering output from Face-GCN, we discard clusters with fewer than 10 samples as a simple heuristic. The remaining cluster assignments on the remaining samples are treated as category labels and merged with the labeled training set. To control for different optimization settings and validation sets, we simply re-train the face recognition model, from scratch, with the same number of epochs and learning rate schedule as the baseline model trained on labeled data -- therefore, the only change between the baseline model and the re-trained model is the extra pseudo-labeled training data.


{\footnotesize
\bibliographystyle{splncs04}
\bibliography{egbib_short}
}